%% file: Main_Arxiv.tex
\documentclass{egpubl}
\usepackage{egsgp2024}

\usepackage{bm}
\usepackage[ruled]{algorithm2e}
\usepackage{xspace}
\usepackage{amssymb}
\usepackage{amsmath}
\usepackage{appendix}
\usepackage{float}
\usepackage{booktabs} 
\usepackage{wrapfig}
\usepackage{color}
\usepackage{diagbox}
\usepackage{makecell}
\usepackage{bbm}
\usepackage{dsfont}
\usepackage{subfigure} 
\usepackage{arydshln}
\usepackage[percent]{overpic}
\usepackage{multirow}
\usepackage{subcaption}
\usepackage{booktabs}
\usepackage{float}
\usepackage{caption}
\usepackage{subcaption}

\SpecialIssueSubmission    

\CGFccby

\usepackage[T1]{fontenc}
\usepackage{dfadobe}  

\usepackage{cite}  
\BibtexOrBiblatex
\electronicVersion
\PrintedOrElectronic

\usepackage{egweblnk}

\newcommand{\ZY}[1]{{\color{black}#1}}

\definecolor{toptimecolor}{RGB}{0,0,255}
\definecolor{topaccolor}{RGB}{240,123,63}
\newcommand{\toptime}[1]{%
    \textcolor{toptimecolor}{\bm{#1}}
}
\newcommand{\topac}[1]{%
    \textcolor{topaccolor}{\bm{#1}}
}

\title{Coverage Axis++: Efficient Inner Point Selection for 3D Shape Skeletonization}

\author[Z. Wang*, Z. Dou* et al.]
{
\parbox{\textwidth}{\centering Zimeng Wang*$^{1}$\orcid{},
Zhiyang Dou*$^{1,2}$\orcid{0000-0003-0186-8269},
Rui Xu$^3$\orcid{0000-0001-8273-1808},
Cheng Lin$^1$\orcid{0000-0002-3335-6623},
Yuan Liu$^1$\orcid{0000-0003-2933-5667},
Xiaoxiao Long$^1$\orcid{0000-0002-3386-8805},\\
\vspace{1mm}
Shiqing Xin$^3$\orcid{0000-0001-8452-8723},
Taku Komura$^1$\orcid{0000-0002-2729-5860},
Xiaoming Yuan$^\dag$$^1$\orcid{},
Wenping Wang$^\dag$$^4$\orcid{0000-0002-2284-3952}   }
        \\
        \vspace{1mm}
{\parbox{\textwidth}{\centering 
$^1$The University of Hong Kong\quad$^2$TransGP\quad$^3$Shandong University\quad$^4$Texas A\&M University  \\
\vspace{1mm}
         \textit{*Equal Contribution}; \textit{$^\dag$Equal Advising}.\\
       }
}
}

\begin{document}

\teaser{
   \centering
 \vspace{-14mm}
\includegraphics[width=\textwidth]{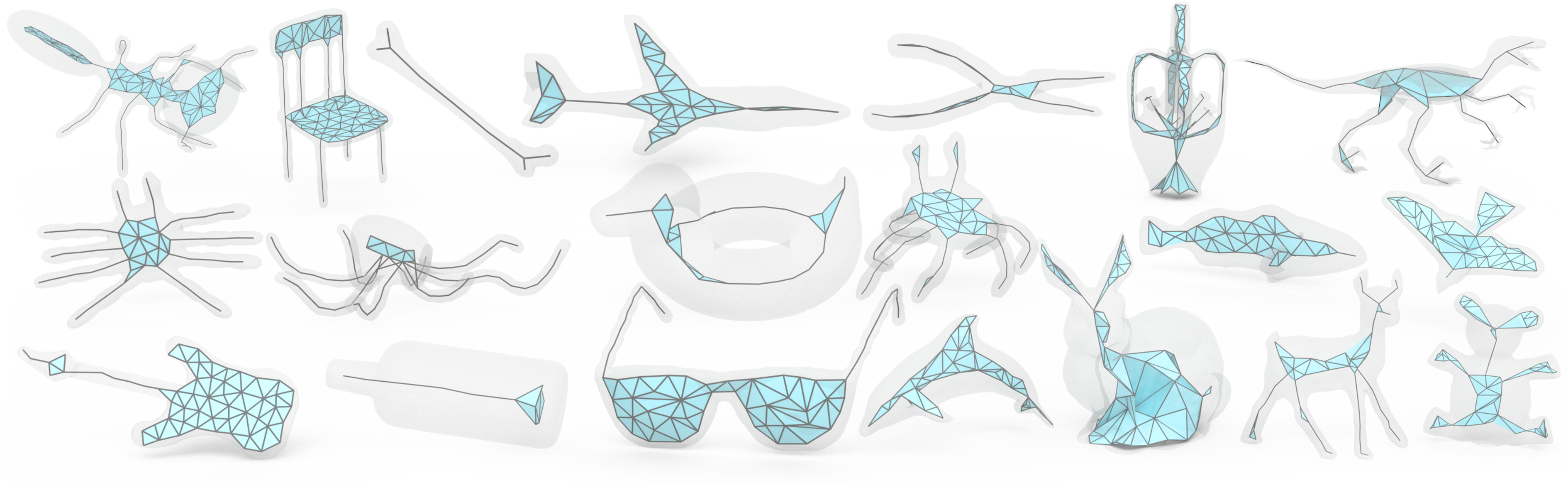} 
\vspace{-7mm}
  \caption{Coverage Axis++ enables highly efficient skeletonization of the given shapes represented by water-tight meshes, triangle soups or point clouds. Compared with the SOTA method~\cite{dou2022coverage, li2015q, pan2019q, lin2021point2skeleton, clemot2023neural}, Coverage Axis++ allows for fast computation of compact medial axis approximation with low reconstruction error, i.e., Coverage Axis++: $5.362\%$ reconstruction error in $2.9$s v.s. Coverage Axis: $6.574\%$ reconstruction error in $236.1$s on average; See Tab.~\ref{table:QMAT}.}
  \label{fig:teaser}
  \vspace{1mm}
}

\maketitle
\begin{abstract}
     We introduce Coverage Axis++, a novel and efficient approach to 3D shape skeletonization. The current state-of-the-art approaches for this task often rely on the watertightness of the input~\cite{li2015q,pan2019q,pan2019q} or suffer from substantial computational costs~\cite{dou2022coverage, clemot2023neural}, thereby limiting their practicality. To address this challenge, Coverage Axis++ proposes a heuristic algorithm to select skeletal points, offering a high-accuracy approximation of the Medial Axis Transform (MAT) while significantly mitigating computational intensity for various shape representations. We introduce a simple yet effective strategy that considers shape coverage, uniformity, and centrality to derive skeletal points. The selection procedure enforces consistency with the shape structure while favoring the dominant medial balls, which thus introduces a compact underlying shape representation in terms of MAT. As a result, Coverage Axis++ allows for skeletonization for various shape representations (e.g., water-tight meshes, triangle soups, point clouds), specification of the number of skeletal points, few hyperparameters, and highly efficient computation with improved reconstruction accuracy. Extensive experiments across a wide range of 3D shapes validate the efficiency and effectiveness of Coverage Axis++. \ZY{Our codes are available at \url{https://github.com/Frank-ZY-Dou/Coverage_Axis}.}

\vspace{-3mm}
\end{abstract}  

\input{Main/1-intro}
\input{Main/2-related}
\input{Main/2.1-Preliminaries}

\input{Main/3-method}

\input{Main/4-exp}

\input{Main/5-conclusion}
\section{Acknowledgements}
\ZY{The authors would like to thank Baorong Yang, Mengyuan Ge, and Ningna Wang for the fruitful discussion and the anonymous reviewers for their valuable comments and suggestions. Zhiyang Dou is supported by the Innovation and Technology Commission of the HKSAR Government under the InnoHK initiative. Shiqing Xin is supported by the National Natural Science Foundation of China (Ref: 62272277). This research is partly supported by Innovation and Technology Commission (Ref: ITS/319/21FP) and Research Grant Council (Ref: 17210222), Hong Kong.
}

\clearpage
\input{Main/6-appendix}

\clearpage
\bibliographystyle{eg-alpha}
\bibliography{egbibsample.bib}

\end{document}

%% file: Main/1-intro.tex
\section{Introduction}
Skeletal representations have become a popular tool in various applications of shape analysis and geometric processing, as they efficiently capture the underlying structures of 3D shapes. They have been widely adopted for various tasks including 3D reconstruction~\cite{wu2015deep, tang2019skeleton, amenta2001power}, volume approximation~\cite{stolpner2011medial, sun2013medial}, shape segmentation~\cite{lin2020seg}, shape abstraction~\cite{dou2020top}, pose estimation~\cite{shotton2011real,yang2021learning, dou2023tore, li2021hybrik}, and animation~\cite{baran2007automatic, yang2018dmat}, among others. With the development of deep learning, skeletal representations also facilitate learning-based approaches in various tasks~\cite{hu2019mat,rebain2021deep,tang2021skeletonnet,hu2022immat,petrov2024gem3d,wen2023learnable,xiong2024speal,wen2024glskeleton,guo2024tetsphere}.

Previous efforts have been made for the computation of curve skeleton~\cite{au2008skeleton, ma2003skeleton, tagliasacchi2012mean, xu2019predicting}, which consists of only 1D curves. Dey and Sun~\cite{dey2006defining} provide a mathematical formulation based on the Medial Geodesic Function, yet curve skeletons are predominantly comprehended empirically.

Medial Axis Transform~(MAT)~\cite{blum1967transformation}, as a complete shape descriptor, is another popular skeletal representation. The MAT is defined by a union of maximally inscribed balls within the shape, accompanied by their respective radius functions. A formal definition of MAT can be found in Sec.~\ref{sec:mat}. In contrast to curve skeletons, the MAT encompasses both curve-like and surface-like structures, offering a consistent definition for arbitrary shapes and superior representational capability.

The MAT computation poses challenges due to sensitivity to boundary noise~\cite{li2015q} and strict input geometry requirements, e.g., watertightness and manifoldness of the surface~\cite{li2015q, pan2019q, wang2022computing}. To tackle the problem, Coverage Axis~\cite{dou2022coverage} is introduced to model skeletal point selection as a Set Cover Problem (SCP), which does not rely on surface connectivity and is able to handle both mesh or point cloud inputs. It aims to identify the smallest sub-collection covering the entire shape. This method minimizes dilated inner balls to approximate the shape, leading to a compact representation of the original shape. However, solving the SCP problem, which is \ZY{known to be} an NP-hard problem, can be time-consuming in some cases. For instance, the average running time of the Coverage Axis can be $236.1$s with the standard deviation being $408.4$s; See Tab.~\ref{table:QMAT}. Meanwhile, the recent work Neural Skeleton~\cite{clemot2023neural} proposes to first learn a Signed Distance Field from the given shape and then use the Coverage Axis to get the final compact skeleton of the given mesh or point cloud. However, it still suffers from low runtime performance (59.27s compared to our 1.63s) and relatively low reconstruction accuracy as shown in Tab.~\ref{table:pc}.

In this paper, we develop a simple yet more efficient skeletonization method for shapes represented as meshes or point clouds. We take shape representation capability, centrality, and uniformity into consideration and develop a heuristic algorithm to achieve representative skeletal point selection with high efficiency. Specifically, we present a scoring scheme that assigns a score to each inner point as a quantification of its skeletal representation ability. The scoring consists of the \textit{coverage}, \textit{uniformity} and \textit{centrality} scores. The coverage score is determined by evaluating the number of surface samples that can be covered within the dilated ball associated with the candidate to ensure coverage inspired by ~\cite{dou2022coverage}. The coverage score ensures consistency with the shape structure by favoring interior points that dominate a larger area, aligning with the definition of the Medial Axis Transform (MAT) as the set of centers of maximally inscribed spheres. On the other hand, uniformity is one of the key desired features of compact skeletonization to suppress redundant structures, for which we introduce the uniformity score. The score is calculated by measuring the distance between the candidate and the nearest point in the set of selected inner points. This achieves a uniformly distributed skeletonization result by preventing the selected points from clustering together. Moreover, the centrality score is employed to encourage those central points to receive higher scores to promote centrality further. The skeletal points are derived by running a priority queue defined by an integration of these scores. We show that this simple yet effective point selection manner produces a better abstraction of the overall shape as well as a shape-aware point distribution without the need to solve an optimization problem with high computation complexity. Notably, although the priority queue works in a greedy manner, Coverage Axis++ achieves global consistency between the skeleton and the original shape and allows for specifying different numbers of target skeletal points; See Sec.~\ref{sec:exp_num_skeleton}. 

Compared with the existing methods~\cite{amenta2001power, li2015q, giesen2009scale, foskey2003efficient, dey2002approximate, sud2007homotopy,miklos2010discrete, dou2022coverage}, Coverage Axis++ surpasses all existing methods in terms of both MAT approximation accuracy and computation efficiency while being able to handle various inputs, including mesh, point cloud or polygon soup. 
To summarize, our method allows for 1) skeletonization for various shape representations including water-tight meshes, triangle soups and point clouds;  2) 
specifying the number of skeletal points, which is a desirable feature for learning-based methods as a fixed number of points are typically required~\cite{hu2019mat, tang2021skeletonnet, hu2022immat,petrov2024gem3d}; 3) highly efficient computation while achieving better or more competitive reconstruction accuracy; 4) few hyperparameters, i.e., only dilation factor is required; 5) randomly generated candidates inside the volume as Coverage Axis++ selects the most expressive ones from overfilled inner point candidates.

We conduct extensive experiments to demonstrate the effectiveness and robustness of our method on a variety of 3D shapes. The comprehensive evaluation reveals that our method effectively captures the underlying MATs with lower shape approximation errors while achieving much higher computation efficiency.

%% file: Main/2-related.tex
\section{Related Work}
Efficient shape representation is essential for various applications of geometric modeling, shape analysis and shape generation~\cite{marschner2023constructive, liu2023surface, sellan2020opening, xu2022rfeps, xu2023globally, dou2020top, fu2022easyvrmodeling,wen2023learnable, lin2020seg, lin2023patch, yang2023neural, zhang2023surface, yariv2023mosaic, yu2023surf, alexa2021polycover, petrov2024gem3d}. Skeletonization, as a popular choice in this regard, has been extensively studied in recent decades.  We refer readers to \cite{tagliasacchi20163d} for a detailed survey covering various forms of skeletons. In the following, we mainly review the related works in Medial Axis Transform~(MAT) computation for mesh and point cloud inputs.

\noindent
\textbf{Medial Axis Transform}  \\
\noindent
\textit{Mesh Inputs.} Efforts in MAT computation given the watertight mesh inputs have been extensive~\cite{amenta2001power, guo2023medial, foskey2003efficient, dey2002approximate, sud2007homotopy, hu2023s3ds, ge2023point2mm, hu2022immat, dou2022coverage, rebain2021deep, wang2022computing,liu2024part123,lin2020modeling}. Specifically, $\lambda$-medial axis~\cite{chazal2005lambda} define weak feature size as a pruning criterion but struggle with feature preservation at different scales~\cite{attali2009stability}. Scale Axis Transform~\cite{miklos2010discrete} excels in spike pruning but is limited by high computational cost and topology disruption. Progressive medial axis~\cite{faraj2013progressive} proposes to perform topology-preserved edge collapse based on sphere absorption. Many studies have been conducted for MAT simplification using different metrics~\cite{marie2016delta,shen2011skeleton,yan2016erosion,yan2018voxel,sun2013medial, li2015q, pan2019q}. Li et al.~\cite{li2015q} propose Q-MAT for highly accurate shape approximation, utilizing a quadratic error metric~\cite{garland1997surface} during simplification. Despite relatively high approximation accuracy, Q-MAT relies on watertight surfaces. It is sensitive to the quality of MAT initialization, as revealed by ~\cite{dou2022coverage}. Later, Q-MAT+~\cite{pan2019q}, as a variant of Q-MAT, further introduces more mesh information like shape diameter function to improve the approximation accuracy and robustness of the original Q-MAT method, especially for those sharp features. Recently, MATFP~\cite{wang2022computing} is developed to compute MAT with well-preserved geometric features via the Restricted Power Diagram for CAD mesh models. Wang et al.~\cite{wang2024mattopo} have also explored addressing the topology preservation issue by employing Restricted Power Diagrams.\\
\noindent
\textit{Point Cloud Inputs.} Point cloud skeletonization is still a fascinating yet challenging problem~\cite{tagliasacchi_sig09, cao2010point, livesu2012reconstructing, li2023epcs, huang2013l1, rebain2019lsmat, wu2015deep, yang2020p2mat,dou2022coverage} due to the absence of manifold information regarding the underlying surface of the shape. Specifically, L1-medial skeleton~\cite{huang2013l1} employs Locally Optimal Projection~\cite{lipman2007parameterization} to contract point clouds to form the skeleton. LSMAT\cite{rebain2019lsmat} approximates the MAT based on the Signed Distance Function (SDF) from a densely sampled oriented point set. Wu et al.\cite{wu2015deep} considers linking surface points to the skeletal points on the meso-skeleton\cite{tagliasacchi2012mean}. Yet, this representation only yields unstructured point sets, which lacks topological constraints describing the skeleton connectivity and leads to low approximation accuracy. Coverage Axis~\cite{dou2022coverage} models skeletal point selection as a Set Cover Problem (SCP), which does not rely on surface connectivity, thus allowing for point cloud inputs. However, it suffers from high computational costs as SCP is NP-hard, especially when facing shapes with planar structures (See detailed analysis in Appendix~D).

With the advent of deep learning, learning-based approaches have been explored for point clouds skeletonization~\cite{yang2020p2mat,lin2021point2skeleton,ge2023point2mm}. For example, Point2Skeleton~\cite{lin2021point2skeleton} learns to predict a set of skeletal points by learning a geometric transformation and then analyzes the connectivity of the skeletal points to form mesh structures from point clouds. However, it suffers from limited generalization due to dependency on training data and lacks accuracy for geometric features. Recently, Neural Skeleton~\cite{clemot2023neural} leverages Implicit Neural Representations (INR), i.e., SDF, together with coverage axis~\cite{dou2022coverage} for skeleton computation, yet it still faces challenges with long running time and dependence on consistently oriented normals for input shapes.

%% file: Main/2.1-Preliminaries.tex
\section{Preliminaries}
\subsection{Medial Axis Transform}
\label{sec:mat}
\begin{wrapfigure}{r}{2.6cm}
\vspace{-4mm}
  \hspace*{-3mm}
  \centerline{
  \includegraphics[width=35mm]{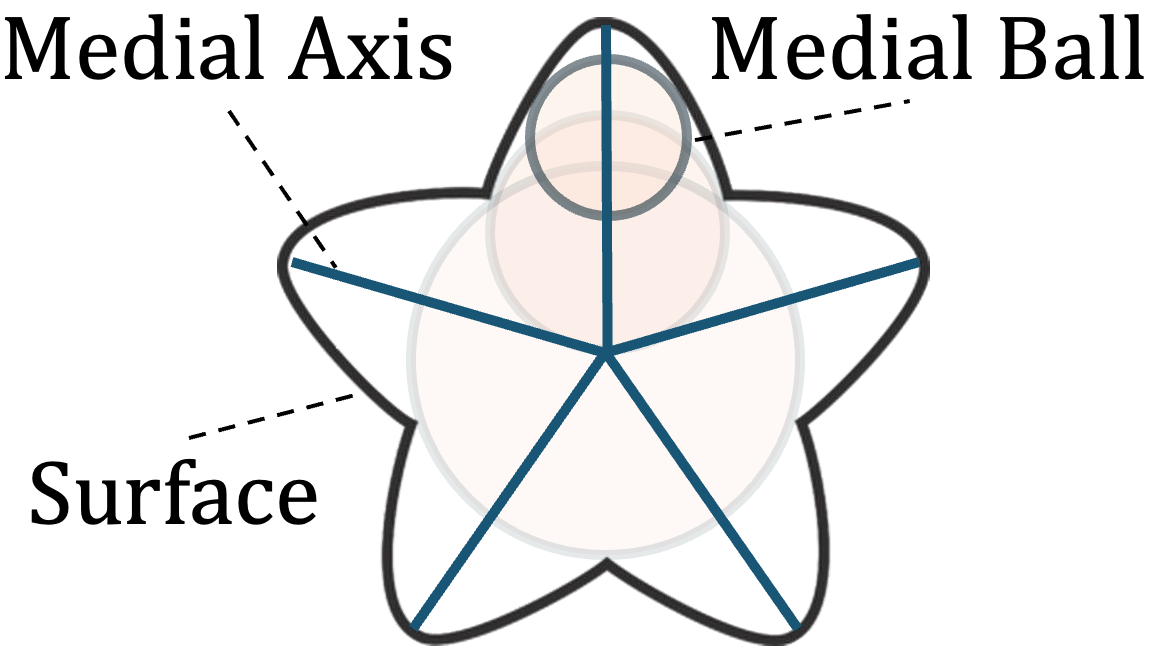}}
  \vspace*{-4mm}
\end{wrapfigure}
For a closed, oriented, and bounded two-manifold surface $\mathcal{S}$ in $\mathbb{R}^3$, the \textit{medial axis} is defined as the locus of the centers of \textit{maximally inscribed spheres} (medial balls) that are tangent to $\mathcal{S}$ at two or more points. Denoted by a pair $\big(\mathcal{M}, \mathcal{R}\big)$, where $\mathcal{M}$ is the medial axis of $\mathcal{S}$ and $\mathcal{R}$ is its radius function, the combination forms the \textit{Medial Axis Transform} (MAT).

\begin{figure*}
\centering
\begin{overpic}
[width=0.98\linewidth]{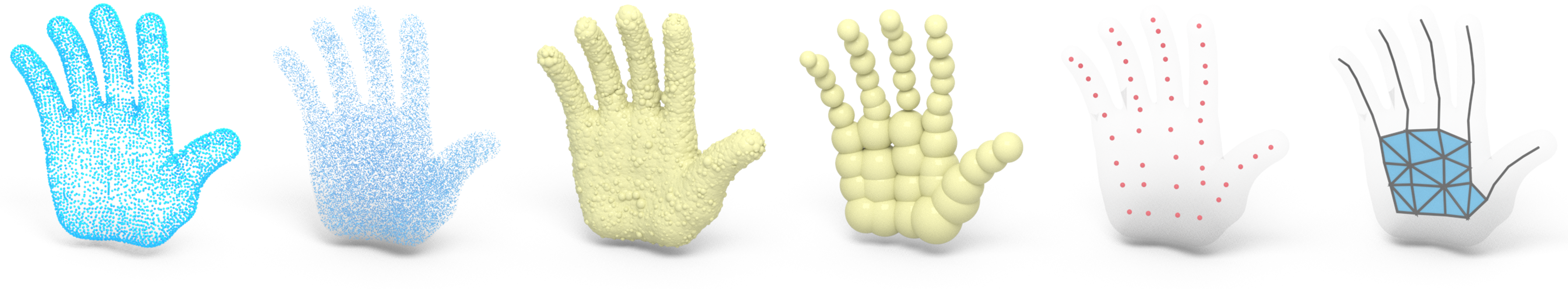}
\put(6.5 ,-0.5){(a)}
\put(23  ,-0.5){(b)}
\put(40  ,-0.5){(c)}
\put(56.5,-0.5){(d)}
\put(73.5,-0.5){(e)}
\put(90.5,-0.5){(f)}
\end{overpic}
\caption{The pipeline of Coverage Axis++. (a) Input point cloud. (b) Inner point candidates. (c) Candidate inner balls with original radius. (d) Selected inner balls with original radius. (e) Selected inner points. (f) Skeletonization result.}
\vspace{-3mm}
\label{fig:pipeline}
\end{figure*}

\subsection{Voronoi Diagram \& Power Diagram} 

Voronoi Diagram (VD) is a partition of the domain $\Omega\subset \mathbb{R}^d$ into regions, close to a set of points called generators $\{\mathbf{x}_i\in\Omega\}_{i=1}^n$. 
Each region, also called cell, is defined as 
\begin{equation}
\Omega_i^{\text{vor}}:~\{\mathbf{x}\in\Omega\;\big|\;\|\mathbf{x}-\mathbf{x}_i\|\leq\|\mathbf{x}-\mathbf{x}_j\|,j\neq i\}.
\nonumber
\end{equation}
A well-known MAT initialization technique is to use the VD inside a shape generated by surface samples~\cite{amenta2001power, li2015q, dou2022coverage}.

In Power Diagram (PD) \cite{aurenhammer1987power}, each generator $\mathbf{x}_i$ is equipped with a weight $w_i$ to control its influence. By defining the power distance $d^{\text{pow}}(x,x_i)$ between $x$ and the weighted generator $x_i$ to be $\|\mathbf{x}-\mathbf{x}_i\|^2-w_i$, the cell associated with $\mathbf{x}_i$ is defined by
\begin{equation}
\Omega_i^{\text{pow}}:~\{\mathbf{x}\in\Omega\;\big|\;d^{\text{pow}}(x,x_i)\leq d^{\text{pow}}(x,x_j),j\neq i\}.
\nonumber
\end{equation}
Therefore, a generator with a larger weight is more dominant.

%% file: Main/3-method.tex
\section{Method}
\label{sec: method}
Our goal is to select skeletal points from a set of candidate points denoted as $P = \{p_i\}$, by utilizing a group of sampled points represented as $S=\{s_j\}$ on the surface, where each point is defined by its 3D Cartesian coordinates in $\mathbb{R}^3$. We present an effective and efficient algorithm for skeletal point selection. We denote by $P^+$ the set of selected skeletal points and $P^-$ the set of unselected ones, such that $P = P^+ \cup P^-$ always holds. We gradually select \ZY{a} point from $P^-$ and add it to $P^+$ until the selected points reach the number specified by the user. In this paper, all inner point candidates are randomly generated inside the volume; See an example in Fig.~\ref{fig:pipeline} (b).
The algorithm for shape skeletonization is outlined through the following steps: scoring each candidate point for point selection and establishing connectivity among skeletal points.

\subsection{Coverage Score}
To achieve compact skeletonization, we aim to select representative points that capture the key features of the given 3D object by measuring its shape coverage~\cite{dou2022coverage} capability.

We first compute the medial radii $R = \{r_i\}$ for the candidate points, where $r_i$ is defined as the closest distance of $p_i$ to the sampled surface set $S$, i.e.,
\begin{equation}\label{eq: radius_ori}
   r_i :=  dist(p_i, S) = \min_{s_j\in S}\|p_i - s_j\|.
\end{equation}
The set $R$ together with the center points $P$ define the candidate balls. We slightly dilate all the balls by a small factor to their radii by offset, namely, 
\begin{equation}
    r_i' = r_i + \delta_r,
    \label{eq:catplus_dilation}
\end{equation}
leading to a set of dilated balls $B= \{B_i\}$, where $B_i$ is the ball centered at $p_i$ with radius $r_i'$. A matrix $\mathbf{D}\in \{0,1\}^{m\times n}$ is introduced by~\cite{dou2022coverage}, where $m$ and $n$ are total numbers of sampled surface points and candidate skeletal points, respectively. Each entry $d_{ji}\in\{0,1\}$ of $\mathbf{D}$ indicates if the surface point $s_j$ is covered by the dilated ball $(p_i, r_i')$:
\begin{equation}
d_{ji}=\left\{
\begin{aligned}
1, & \ \text{if} \ \left\|p_i - s_j\right\|_2 \le r_i', j = 1,...,m, \ i = 1, ... , n\\
0, & \ \text{if} \ \left\|p_i - s_j\right\|_2 > r_i', j = 1,...,m, \ i = 1, ... , n.  \\
\end{aligned}
\right.
\label{eq:coverage_matrix}
\end{equation}

According to Eq.~\ref{eq:coverage_matrix}, we have the coverage matrix $\mathbf{D} = (d_{ji})\in \mathbb{R}^{m \times n}$, where $d_{ji} \in \{0, 1\}$ takes the value $1$ if the surface point $s_j$ is covered by the dilated ball $B_i$ and $0$ otherwise. 

In this paper, we define $S'$ as the uncovered surface points, i.e.,
\begin{equation} \label{eq:S'}
    S' :~\{s_j\in S\;\big|\;s_j\notin B_i,\;\forall p_i\in P^+\}.
\end{equation}
Then we present the \textit{coverage score} to each $p_i\in P^-$, which is determined by the number of samples in uncovered surface points $S'$ that $B_i$ covers
\begin{equation}
    Cov_i := \sum_{j: s_j\in S'} d_{ji}.
    \label{eq:cover_score}
\end{equation}
If the dilated ball $B_i$ of a candidate point $p_i$ can cover lots of surface samples from the surface point set $S'$, $p_i$ should be regarded as a representative candidate, which indicates that it has high (local) shape approximation capability. The coverage score favors the medial balls that dominate a larger local volume of the given shape during the selection and thus promotes a high coverage rate (See Appendix.~C) of the overall shape to effectively enforce consistency with the input shape.

\subsection{Uniformity Score}
In addition to the coverage score that intrinsically relates the skeletal points with the model surface, one needs to guarantee uniformity among the selected skeletal points.
In other words, the selected points themselves should be relatively uniformly distributed in the space instead of clustering together to avoid redundancy. For this purpose, we introduce the \textit{uniformity score}, which aims at punishing those candidate points close to the set of currently selected points $P^+$. Specifically, for each $p_i\in P^-$, its uniformity score is defined as
\begin{equation}\label{eq:uniform}
    Unif_i := dist(p_i, P^+) = \min_{p_{i'}\in P^+}\|p_i - p_{i'}\|.
\end{equation}
\begin{wrapfigure}{r}{2cm}
\vspace{-3.5mm}
  \hspace*{-5mm}
  \centerline{
  \includegraphics[width=25mm]{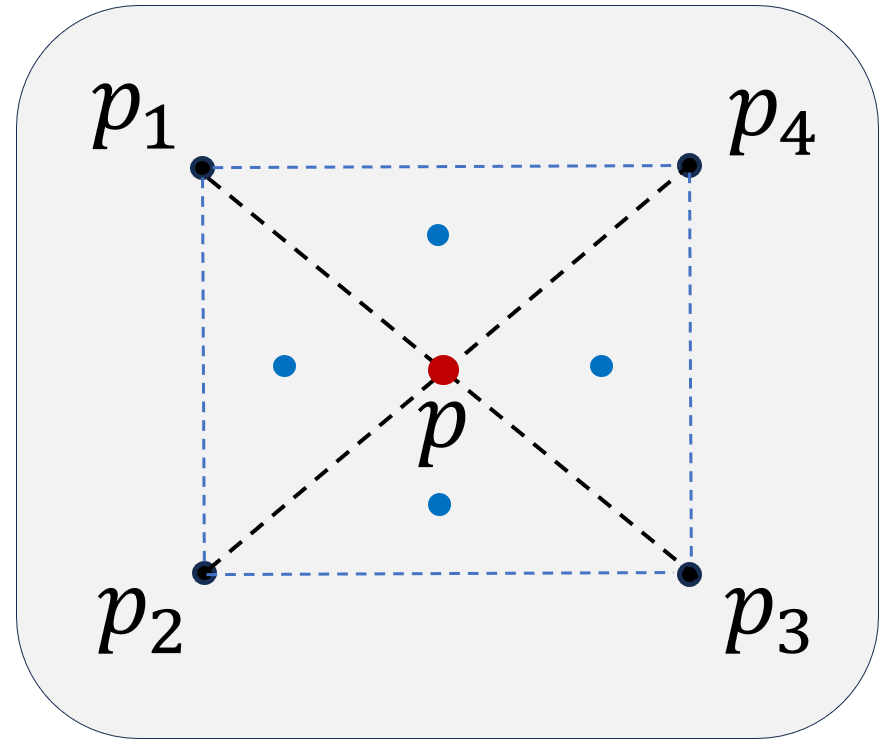}}
  \vspace*{-4mm}
\end{wrapfigure} 
To illustrate the mechanics of the uniformity score defined above, consider skeletal point selection for a simple planar shape, where the $p_1, p_2, p_3, p_4$ are selected points within a rectangular shape. If not considering the effect of the coverage score, then according to Eq.~\ref{eq:uniform}, the next skeletal point $p$ (highlighted in red) should be the intersection point of the line segments connecting $p_1$ and $p_3$, as well as $p_2$ and $p_4$, ensuring that $\min_{i\in\{1, 2, 3, 4\}}\|p-p_i\|$ is the largest within the rectangle, leading to the highest possible uniformity score for the selected point.
Following this pattern, the subsequent four selected points (highlighted in blue) should be placed at the centroids of their respective triangles, resulting in a relatively uniform distribution of the selected skeletal points.

\subsection{Centrality Score}
We further encourage the centrality of the selected points by introducing a centrality score for each candidate skeletal point. The more central a point is in terms of medial ball positioning, the higher its score. It excludes points too close to the surface to avoid their susceptibility to surface irregularities and noise.
Specifically, for each $p_i\in P^-$, we define its centrality score based on its medial radius by
\begin{equation}\label{eq:reg}
    Centr_i := -\frac{1}{r_i}, 
\end{equation}
where $r_i$ is the radius of $p_i$ defined in Eq.~\ref{eq: radius_ori}.

\subsection{Final Score}
We assign a final score to each candidate $p_i\in P^-$ by combining its coverage score, uniformity score, and centrality score.
We denote the three score vectors as $\mathbf{Cov} = (Cov_i)$, $\mathbf{Unif} = (Unif_i)$, and $\mathbf{Centr} = (Centr_i)$, respectively.
Since these scores have different scales, we apply standardization before integrating them. Specifically, for each $p_i \in P^-$, we perform the following transformation to $Cov_i$, $Unif_i$ and $Centr_i$
\begin{equation}\label{eq:standardization}
\begin{aligned}
     & Cov_i \leftarrow \frac{Cov_i - mean(\mathbf{Cov})}{std(\mathbf{Cov})},\\
     & Unif_i \leftarrow \frac{Unif_i - mean(\mathbf{Unif})}{std(\mathbf{Unif})},\\
     & Centr_i \leftarrow \frac{Centr_i - mean(\mathbf{Centr})}{std(\mathbf{Centr})},
\end{aligned}
\end{equation}
where $mean(\cdot)$ and $std(\cdot)$ represent the sample mean and sample standard deviation respectively.
Then we integrate the three scores by summing them up, i.e.,
\begin{equation}\label{eq:score}
    Score_i := Cov_i + Unif_i + Centr_i.
\end{equation}
After computing $Score_i$ for all $p_i\in P^-$, we select the candidate that achieves the highest score.
An overview of the heuristic algorithm is summarized in Algorithm \ref{heuristic}. An ablation study on different scoring schemes is given in Sec.~\ref{subsubsec:scoring}.

\begin{algorithm}
  \SetAlgoLined
  \KwIn{sampled surface points $S$, candidate skeletal points $P$, dilation factor $\delta_r$, target number of selected points $|V|$.}
  \KwResult{the set of selected skeletal points $P^+$}
  Compute matrix $\mathbf{D}$ based on Eq.~\ref{eq:coverage_matrix}, 
  Initialize $S' = S$, $P^+ = \emptyset$, $P^- = P$ and $k=1$\;
  \While{$S'\neq \emptyset$ and $k\leq |V|$}{
    \For{$p_i\in P^-$}{
    Compute $Cov_i$ according to Eq.~\ref{eq:cover_score}\;
    Compute $Unif_{i}$ according to Eq.~\ref{eq:uniform}\;
    Compute $Centr_{i}$ according to Eq.~\ref{eq:reg}\;
    }
    Compute the mean and standard deviation for the two vectors $\mathbf{Cov} = (Cov_i)$ and $\mathbf{Unif} = (Unif_i)$\;
    \For{$p_i\in P^-$}{
    Standardize $Cov_i$, $Unif_i$ and $Centr_i$ according to Eq.~\ref{eq:standardization}\;
    Compute $Score_i$ according to Eq.~\ref{eq:score}\;
    }
    Select $p_{i_k} \leftarrow \arg\max_{p_i\in P^-} \{Score_i\}$\;
    Update $P^+ \leftarrow P^+\cup \{p_{i_k}\}$, $P^- \leftarrow P^-\setminus \{p_{i_k}\}$\;
    Update $S'$ according to Eq.~\ref{eq:S'}\;
    Set $k \leftarrow k+1$\;
    }
  \caption{Coverage Axis++ for inner point selection}\label{heuristic}
\end{algorithm}

\subsection{Candidate and Surface Point Generation}
\label{sec:point_generation_sec}
In this paper, for mesh or point cloud inputs, we always randomly generate the set of candidate points $P$ ($|P|$ = 10000) inside the volume using volumetric sampling\footnote{https://www.meshlab.net}. The number of surface samples $S$ is $1500$ in all the comparisons (except for Appendix.~A). For mesh input, we employ Poisson-disk sampling~\cite{corsini2012efficient} to get the surface sample. For point cloud input, surface samples are obtained by white noise sampling~\cite{gptoolbox}, and fast winding number~\cite{barill2018fast} is used for inside-outside determination.

\subsection{Connectivity of Skeletal Points}
\label{subsec:connection}

\subsubsection{Mesh Input} 
\label{sec:mesh_input_connection}

\begin{figure}[t]
    \centering
  \begin{overpic}
[width=\linewidth]{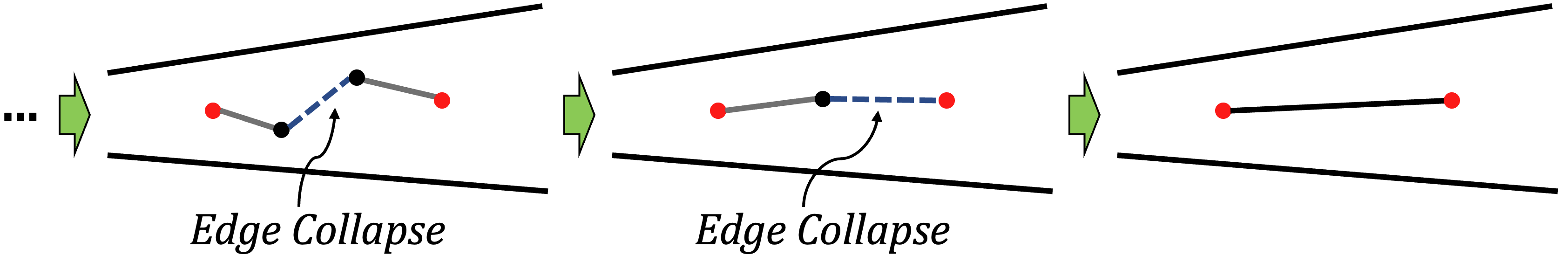}
\put(18,-3){(a)}
\put(51,-3){(b)}
\put(84,-3){(c)}
\end{overpic}
 \vspace{-3mm}
    \caption{\ZY{An illustration of the edge collapse process. The surface is in black, with selected skeletal points highlighted in red. The collapsed edges are shown with dashed lines.}}
    \label{fig:edge_collapse}
    \vspace{-6mm}
\end{figure}

For a mesh input, we aim to build up a connection while preserving the topology of the skeleton, for which we employ the edge collapse strategy of Q-MAT \cite{li2015q} which is based on a Voronoi diagram. Following~\cite{amenta2001power, wang2022restricted, dou2022coverage}, we initialize the initial Voronoi diagram $\textbf{VD}$ using the surface points. We then embed the selected points in the Voronoi diagram using the nearest projection, following~\cite{dou2022coverage}. We simplify $\textbf{VD}$ using the edge collapse strategy~\cite{li2015q}.  \ZY{Specifically, during this process, each edge is checked and collapsed using the edge collapse strategy~\cite{li2015q} based on the Spherical Quadric Error Metric~\cite{li2015q, thiery2013sphere}. We use additional conditions during edge collapse: If both endpoints of an edge to be collapsed are not selected points, we directly apply the edge collapse strategy following~\cite{li2015q}. If one endpoint of an edge is a selected point and the other is not, during edge collapse, the selected point is preserved. If both endpoints of an edge are selected points, the edge is preserved and skipped. This process continues until all edges have endpoints that belong to the selected points; See Fig.~\ref{fig:edge_collapse}.} We adopt topology preservation~\cite{dey1999topology} and mesh inversion avoidance strategy~\cite{garland1997surface} during the collapse. The simplification result $\textbf{VD}'$ serves as the final MAT of the given mesh input. 

\noindent \textit{{Remark.}} Notably, not all Voronoi vertices inside the given shape can be used to approximate the medial surface. One has to use \textit{Pole}s (a subset of inside Voronoi vertices) for medial surface approximation according to~\cite{amenta2001power}. However, in practice, we find the collapse over the original Voronoi diagram $\textbf{VD}$ yields relatively stable outcomes for a compact medial surface approximation. \ZY{Furthermore, the connectivity of skeletal points (edges and facets in MAT) is crucial information in the output MAT results. This connectivity is significantly important for downstream applications such as shape analysis and reconstruction, which we discuss in detail in Appendix~E.}

\begin{wrapfigure}{r}{3cm}
\vspace{-3.5mm}
  \hspace*{-5mm}
  \centerline{
  \includegraphics[width=35mm]{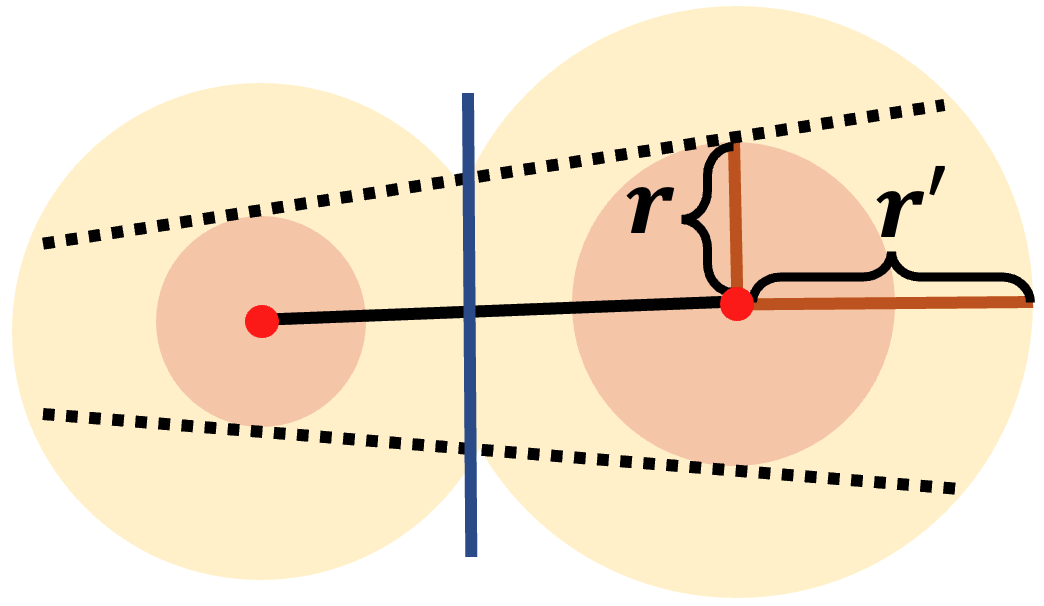}}
  \vspace*{-4mm}
\end{wrapfigure} 
\subsubsection{Point Cloud Input} Topology preservation during connection establishment for point clouds is challenging. The aforementioned edge collapse techniques~\cite{li2015q, pan2019q} are not applicable as they require a watertight surface. Thus, we approximate the MAT connection using the Power Diagram following~\cite{amenta2001power}. We compute the Power diagram $\textbf{PD}$ on the set of $\{B'(p_i,r_i'), p_i \in P^+\} \cup \{B(s_j, \delta_r), s_j \in S\} $ where $P^+$, $S$ and \ZY{$r_i' = r_i + \delta_r$~(Eq.~\ref{eq:catplus_dilation}) are sets of selected points, surface points and dilated radius, respectively}. Then, the connectivity among skeletal points is obtained by extracting edges among the selected inner points on the dual of $\textbf{PD}$, regular triangulation $\textbf{RT}$ (in other words, weighted alpha complexes). \ZY{In the inset figure, the \( P^+ \) points are depicted in red, while the surface samples are shown in black. The medial balls with original radius \( r \) and dilated radius \( r' \) are illustrated in red and yellow, respectively. Connectivity is established by the connections among selected points of the regular triangulation \(\mathbf{RT}\), which are also colored as black solid lines. Additionally, the power plane between the two sites is visualized in blue.} This approach can also be applied to mesh inputs, as the computation of the Power Diagram does not depend on the underlying surface being watertight. An overview of our pipeline is shown in Fig.~\ref{fig:pipeline}.

\subsection{Discussion}
\label{subsec: algo-benifit}
\subsubsection{Complexity} 
Our method (Algorithm \ref{heuristic}) has a polynomial complexity.
Specifically, computing $Cov_i$ and $Centr_i$ costs $O(|S|)$ operations, while $Unif_i$ requires $O(|V|)$ operations. 
Hence summing up for all $p_i\in P^-$ gives complexity $O(|P|\cdot|S|)$, considering that $|V|$ is usually very small compared to $|S|$ and $|P|$. Standardizing $\mathbf{Cov}$, $\mathbf{Unif}$ and $\mathbf{Centr}$, as well as sorting, each take $O(|P|)$ operations. 
Taking into account that the procedure repeats for a maximum of $|V|$ times, the overall complexity of Algorithm \ref{heuristic} can be expressed as $O(|S|\cdot|P|\cdot |V|)$.
Fig.~\ref{fig:time_ablation} validates this complexity empirically by illustrating the impact of different set sizes $|P|$, $|S|$ and $|V|$ on the running time of Coverage Axis++.
In contrast, Coverage Axis requires an exponential cost in the worst case. Although modern solvers accelerate the solving process, long computation time is incurred in the worst case; See Tab.~\ref{table:QMAT}. 

\begin{table*}
\footnotesize
\begin{center}
\vspace{13mm}
\caption{Quantitative comparison on run time and shape approximation error between Coverage Axis and Coverage Axis++. The inputs are meshes and point clouds. \textbf{Time}: \textit{Runtime measured in seconds.} $|V|$: \textit{The number of skeletal points.} We set the number of selected skeletal points of our method the same as the Coverage Axis~\cite{dou2022coverage} for a fair comparison.}
\label{table:QMAT}
\vspace{-1mm}
\resizebox{\textwidth}{!}{
\setlength{\tabcolsep}{0.6mm}{
\begin{tabular}{p{1.5cm}|p{0.9cm}<{\centering}|cccc|cccc|cccc|cccc}
\hline
\multirow{3}{*}{Model} & \multirow{3}{*}{$|V|$ } & \multicolumn{8}{c|}{Mesh}  & \multicolumn{8}{c}{Point Cloud}\\
\cline{3-18}
  &  &  \multicolumn{4}{c|}{Coverage Axis} & \multicolumn{4}{c|}{Coverage Axis++} & \multicolumn{4}{c|}{Coverage Axis} & \multicolumn{4}{c}{Coverage Axis++}  \\
  &  &  \textbf{Time} & $\overrightarrow{\epsilon}$ &  $\overleftarrow{\epsilon}$ &  $\overleftrightarrow{\epsilon}$ & \textbf{Time} &  $\overrightarrow{\epsilon}$ &  $\overleftarrow{\epsilon}$ &  $\overleftrightarrow{\epsilon}$ & \textbf{Time} & $\overrightarrow{\epsilon}$ &  $\overleftarrow{\epsilon}$ &  $\overleftrightarrow{\epsilon}$ & \textbf{Time} & $\overrightarrow{\epsilon}$ &  $\overleftarrow{\epsilon}$ &  $\overleftrightarrow{\epsilon}$\\
    \hline
    Bear-1 & $68$ & $12.6$ & $3.964\%$ & $4.107\%$ & $4.107\%$ & \toptime{$1.7$} & $3.846\%$ & $4.047\%$ & \topac{$4.047\%$} & 14.1	& 4.707\%	& 6.031\%	& 6.031\%	& \toptime{$3.3$}	& 5.092\%	& 6.016\%	& \topac{$6.016\%$}
\\
    Bear-2 & $49$ & $2.3$ & $4.345\%$ & $5.016\%$ & $5.016\%$  & \toptime{$1.0$} &	$4.313\%$ & $4.572\%$ & \topac{$4.572\%$} & 3.6	& 8.220\%	& 8.103\%	& 8.220\%	& \toptime{$2.3$}	& 3.739\%	& 4.531\%	& \topac{$4.531\%$}
 \\
	Bird & $83$ & >1000 & $2.879\%$ & $2.861\%$ & $2.879\%$ & \toptime{$2.3$} & $2.306\%$ & $2.560\%$ & \topac{$2.560\%$} & >1000	& 3.166\%	& 6.810\%	& 6.810\%	& \toptime{$3.9$}	& 2.590\%	& 2.561\%	& \topac{$2.590\%$}
\\ 
    Bug & $137$ &	$14.4$ & $4.632\%$ & $4.586\%$ & $4.632\%$ & \toptime{$5.4$}	& $3.114\%$ &	$4.511\%$ & \topac{$4.511\%$} & 16.2	& 9.208\%	& 8.366\%	& 9.208\%	& \toptime{$7.5$}	& 8.262\%	& 4.398\%	& \topac{$8.262\%$}
\\
    Bunny & $73$ & $2.1$ & $7.625\%$ & $7.634\%$ & $7.634\%$ &\toptime{$1.5$} & $4.860\%$ & $5.771\%$	& \topac{$5.771\%$} & 3.4	& 6.391\%	& 11.398\%	& 11.398\%	& \toptime{$3.0$}	& 4.324\%	& 6.091\%	& \topac{$6.091\%$}
\\
    Camel & $93$ & $38.5$ & $3.487\%$ & $6.420\%$ &	$6.420\%$ &\toptime{$2.5$ }& $2.175\%$ &	$5.556\%$ & \topac{$5.556\%$} & 40.0	& 8.340\%	& 7.552\%	& 8.340\%	& \toptime{$4.1$}	& 3.706\%	& 5.108\%	& \topac{$5.108\%$}
\\
    Chair &	$115$ & >1000 & $2.113\%$ &	$2.119\%$ &  \topac{$1.903\%$} &	\toptime{$4.5$} & $2.035\%$ &	$2.947\%$ & {$2.947\%$}  & >1000 & 3.210\%	& 4.199\%	& 4.199\%	& \toptime{$6.2$}	& 3.160\%	& 2.890\%	& \topac{$3.160\%$}
\\
    Crab-1 & $93$ & $17$ & $2.546\%$ & $2.592\%$ & \topac{$2.592\%$} & \toptime{$2.9$} & $2.744\%$ &	$2.608\%$ & {$2.744\%$} & 18.5	& 3.476\%	& 3.232\%	& 3.476\%	& \toptime{$4.5$}	& 2.862\%	& 3.207\%	& \topac{$3.207\%$}
\\
    Crab-2 & $107$ & $16.4$	& $2.901\%$ &	$2.593\%$ & $2.901\%$ & \toptime{$4.0$} & $2.825\%$ &	$2.767\%$ & \topac{$2.825\%$} & 17.9	& 3.221\%	& 3.878\%	& 3.878\%	& \toptime{$5.7$}	& 2.764\%	& 2.778\%	& \topac{$2.778\%$}
\\
    Cup	& $243$ & >1000 &	$9.840\%$ & $6.031\%$ & $9.840\%$	& \toptime{$13.8$} & $9.247\%$ & $6.251\%$ & \topac{$9.247\%$}  & >1000	& 6.658\%	& 7.214\%	& 7.214\%	& \toptime{$16.4$}	& 6.701\%	& 3.809\%	& \topac{$6.701\%$}
\\
    Dinosaur & $66$ & $2.4$ & $2.903\%$ &	$2.896\%$ & $2.903\%$ &	\toptime{$1.4$} & $2.308\%$ &	$2.672\%$ & \topac{$2.672\%$} & 3.8	& 1.978\%	& 2.175\%	& \topac{$2.175\%$}	& \toptime{$2.8$}	& 2.920\%	& 3.287\%	& 3.287\%
\\
    Dolphin	& $30$ & $14.1$ & $16.926\%$ & $42.999\%$ &	$42.999\%$ & \toptime{$0.7$} & $4.916\%$ & $15.323\%$ & \topac{$15.323\%$} & 15.3	& 16.840\%	& 42.768\%	& 42.768\%	& \toptime{$2.1$}	& 6.652\%	& 21.393\%	& \topac{$21.393\%$}
\\ 
    Elephant & $94$ & $2.9$ & $3.216\%$ & $3.525\%$ & \topac{$3.525\%$} &	\toptime{$2.4$} & $3.373\%$ &	$3.874\%$ & {$3.874\%$} & 4.4	& 3.519\%	& 3.732\%	& \topac{$3.732\%$}	& \toptime{$4.0$}	& 3.395\%	& 6.155\%	& 6.155\%
\\
    Fandisk	& $127$ &	$401.8$ &	$5.012\%$ & $4.597\%$ &	$5.012\%$ & \toptime{$4.6$} &	$3.849\%$ & $4.067\%$ & \topac{$4.067\%$} & 403.3	& 11.310\%	& 9.870\%	& 11.310\%	& \toptime{$6.3$}	& 3.503\%	& 4.298\%	& \topac{$4.298\%$}
\\
    Femur &	$26$ & $2.0$ & $2.288\%$ & $2.272\%$ & $2.288\%$  & \toptime{$0.4$} & $2.314\%$ &	$3.095\%$ & \topac{$3.095\%$} & 3.4	& 4.703\%	& 4.953\%	& 4.953\%	& \toptime{$2.0$}	& 2.231\%	& 3.010\%	& \topac{$3.010\%$}
\\
    Fish & $43$ &	$41.4$ & $3.686\%$ & $9.077\%$ & $9.077\%$ & \toptime{$1.1$} & $3.421\%$ &	$4.809\%$ & \topac{$4.809\%$} & 42.7	& 7.824\%	& 12.980\%	& 12.980\%	& \toptime{$2.5$}	& 3.627\%	& 4.155\%	& \topac{$4.155\%$}
\\
    Giraffe	& $71$ & $2.8$ & $1.914\%$ & $2.396\%$ & \topac{$2.396\%$} & \toptime{$1.7$} & $1.831\%$ &	$4.208\%$ & {$4.208\%$} & 4.3	& 4.389\%	& 3.260\%	& 4.389\%	& \toptime{$3.2$}	& 2.727\%	& 3.975\%	& \topac{$3.975\%$}
\\ 
    Guitar & $71$ & >1000 & $2.255\%$ &	$2.182\%$ & \topac{$2.255\%$} &	\toptime{$1.7$} & $2.136\%$ & $2.497\%$ & {$2.497\%$}  & >1000 & 2.072\%	& 2.276\%	& \topac{$2.276\%$}	& \toptime{$3.3$}	& 4.054\%	& 2.426\%	& 4.054\%
\\
    Hand & $47$ &	$11.2$ & $2.104\%$ & $2.074\%$ & $2.104\%$ & \toptime{$1.1$} & $1.974\%$ & $1.975\%$ & \topac{$1.975\%$} & 12.4	& 8.900\%	& 8.428\%	& 8.900\%	& \toptime{$5.9$}	& 2.237\%	& 2.328\%	& \topac{$2.328\%$}
\\
    Human & $46$ & $2.2$ & $1.827\%$ & $1.906\%$ & \topac{$1.906\%$} & \toptime{$0.9$} & $2.062\%$ &	$2.105\%$ & {$2.105\%$} & 3.4	& 2.425\%	& 3.247\%	& \topac{$3.247\%$}	& \toptime{$2.3$}	& 1.959\%	& 6.961\%	& 6.961\%
\\
    Kitten & $148$ & $5.0$ & $5.804\%$ & $6.679\%$ & \topac{$6.679\%$} &  \toptime{$5.0$} & $5.839\%$ & $7.798\%$ & {$7.798\%$} & \toptime{$6.6$}	& 7.415\%	& 7.941\%	& 7.941\%	& 6.8	& 7.415\%	& 7.941\%	& \topac{$7.941\%$}
\\
    Lifebuoy & $33$ & $1.5$ & $2.856\%$ & $2.251\%$ & $2.856\%$ & \toptime{$0.5$} & $3.324\%$ &	$3.217\%$ & \topac{$3.324\%$} & 2.8	& 5.252\%	& 4.645\%	& 5.252\%	& \toptime{$1.9$}	& 3.700\%	& 3.669\%	& \topac{$3.700\%$}
\\
    Neptune & $53$ & $3.0$ & $5.947\%$ & $11.030\%$ & $11.030\%$ &\toptime{$1.2$} &	$3.227\%$ & $9.989\%$ & \topac{$9.989\%$} & 4.5	& 4.768\%	& 14.161\%	& 14.161\%	& \toptime{$2.8$}	& 5.435\%	& 5.485\%	& \topac{$5.485\%$}
\\
    Octopus & $74$ & \toptime{$0.4$} & $2.556\%$ & $3.339\%$ & $3.339\%$ & $1.8$ & $2.696\%$ &	$3.005\%$ & \topac{$3.005\%$} & \toptime{$1.7$}	& 2.488\%	& 3.231\%	& 3.231\%	& 3.5	& 2.618\%	& 3.214\%	& \topac{$3.214\%$}
\\
    Plane & $44$ & $64.4$ &	$4.660\%$ & $10.032\%$ & $10.032\%$ & \toptime{$1.0$}	& $2.534\%$ &	$4.306\%$ & \topac{$4.306\%$} & 65.9	& 4.594\%	& 17.325\%	& 17.325\%	& \toptime{$2.4$}	& 3.704\%	& 4.500\%	& \topac{$4.500\%$}
\\
    Pot	& $90$ & $79.8$ & $4.376\%$ &	$3.562\%$ & \topac{$4.376\%$} &	\toptime{$2.3$} & $3.979\%$ & $5.278\%$ & {$5.278\%$} & 81.2	& 3.631\%	& 5.043\%	& 5.043\%	& \toptime{$3.6$}	& 3.548\%	& 3.726\%	& \topac{$3.726\%$}
\\
    Rapter & $44$ & $10.6$ & $7.073\%$ & $14.310\%$ & $14.310\%$ &  \toptime{$1.0$} & $4.089\%$ & $10.727\%$ & \topac{$10.727\%$} & 12.0	& 5.062\%	& 18.442\%	& 18.442\%	& \toptime{$2.4$}	& 7.425\%	& 9.307\%	& \topac{$9.307\%$}
\\
    Rocker & $112$ & >1000 & $3.660\%$ & $2.721\%$ & $3.660\%$ & \toptime{$3.5$} & $3.652\%$ & $3.274\%$ & \topac{$3.652\%$} & >1000	& 5.099\%	& 5.736\%	& 5.736\%	& \toptime{$5.1$}	& 3.570\%	& 2.989\%	& \topac{$3.570\%$}
\\
    Seahorse & $60$ & $37.6$ & $2.236\%$ & $3.164\%$ & \topac{$3.164\%$} & \toptime{$1.4$} & $2.296\%$ & $5.168\%$ & {$5.168\%$} & 39.0	& 3.197\%	& 5.592\%	& \topac{$5.592\%$}	& \toptime{$2.9$}	& 2.256\%	& 5.878\%	& 5.878\%
\\
    Spectacles & $97$ & >1000 & $2.238\%$ &	$2.319\%$ & \topac{$2.319\%$}	& \toptime{$2.8$} & $2.323\%$ & $3.080\%$ & {$3.080\%$} & >1000	& 1.984\%	& 2.395\%	& \topac{$2.395\%$}	& \toptime{$4.3$}	& 2.329\%	& 2.921\%	& 2.921\%
\\
    Spider & $54$	& $3.8$ & $4.629\%$ & $5.295\%$ & $5.295\%$ & \toptime{$1.1$} & $2.731\%$ &	$4.493\%$ & \topac{$4.493\%$} & 5.3	& 5.144\%	& 25.817\%	& 25.817\%	& \toptime{$2.8$}	& 4.532\%	& 5.555\%	& \topac{$5.555\%$}
\\
    Vase & $118$ & \toptime{$1.1$} & $3.779\%$ & $5.971\%$ & $5.971\%$ & $3.3$ & $3.575\%$ & $3.875\%$ & \topac{$3.875\%$} & \toptime{$2.9$}	& 4.199\%	& $6.034\%$	& $6.034\%$	& $5.2$	& 4.069\%	& 5.006\%	& \topac{$5.006\%$}
\\
    Wine glass & $278$ & >1000 & $21.310\%$ & $11.724\%$ & \topac{$21.310\%$} & \toptime{$16.7$} & $22.851\%$ & $8.828\%$ & {$22.851\%$} & >1000	& $13.673\%$	& $8.254\%$	& \topac{$13.673\%$}	& \toptime{$19.2$}	& $13.868\%$	& $5.952\%$	& $13.868\%$
\\
\hline
{Average} & {-} & $236.1$ & $4.775\%$ & $6.069\%$ & 6.574\% & \toptime{$2.9$} & $3.902\%$ & $4.826\%$ & \topac{$5.362\%$} & $237.7$ & $5.669\%$ & $8.639\%$ & $8.974\%$ & \toptime{$4.6$} & $4.272\%$ & $5.016\%$ & \topac{$5.537\%$}
 \\
\hline
\end{tabular}
}
}
\end{center}
\vspace{0mm}
\end{table*}

\subsubsection{More Features} 
\label{subsec:more_features}
Coverage Axis++ offers control over the number of selected inner points compared to Coverage Axis. In Sec.~\ref{sec:exp_num_skeleton}, we explore its robustness to various target numbers of selected skeletal points. This ability to specify a specific number of skeletal points is crucial for learning-based methods, ensuring compatibility with algorithms that assume a fixed number of input points~\cite{hu2019mat, tang2021skeletonnet, hu2022immat, petrov2024gem3d}.  Unlike Coverage Axis~\cite{dou2022coverage} or Neural Skeleton~\cite{clemot2023neural}, which enforces a hard constraint requiring all surface samples to be covered, our method relaxes this constraint, enhancing robustness to coverage conditions. This relaxation is particularly advantageous when dealing with a large number of surface samples. Coverage Axis++ maintains a high coverage rate of the original shape, as demonstrated in Appendix.~C.

Coverage Axis++ does not impose strict requirements on the watertightness or manifoldness of the input compared with methods like Q-MAT~\cite{li2015q, pan2019q}, making it versatile for handling meshes and point clouds.

Additionally, our method only requires two hyperparameters: target skeletal number $|V|$, and dilation factor $\delta_r$.

%% file: Main/4-exp.tex
\begin{figure*}[]
\vspace{10mm}
\centering
\begin{overpic}[width=\linewidth]{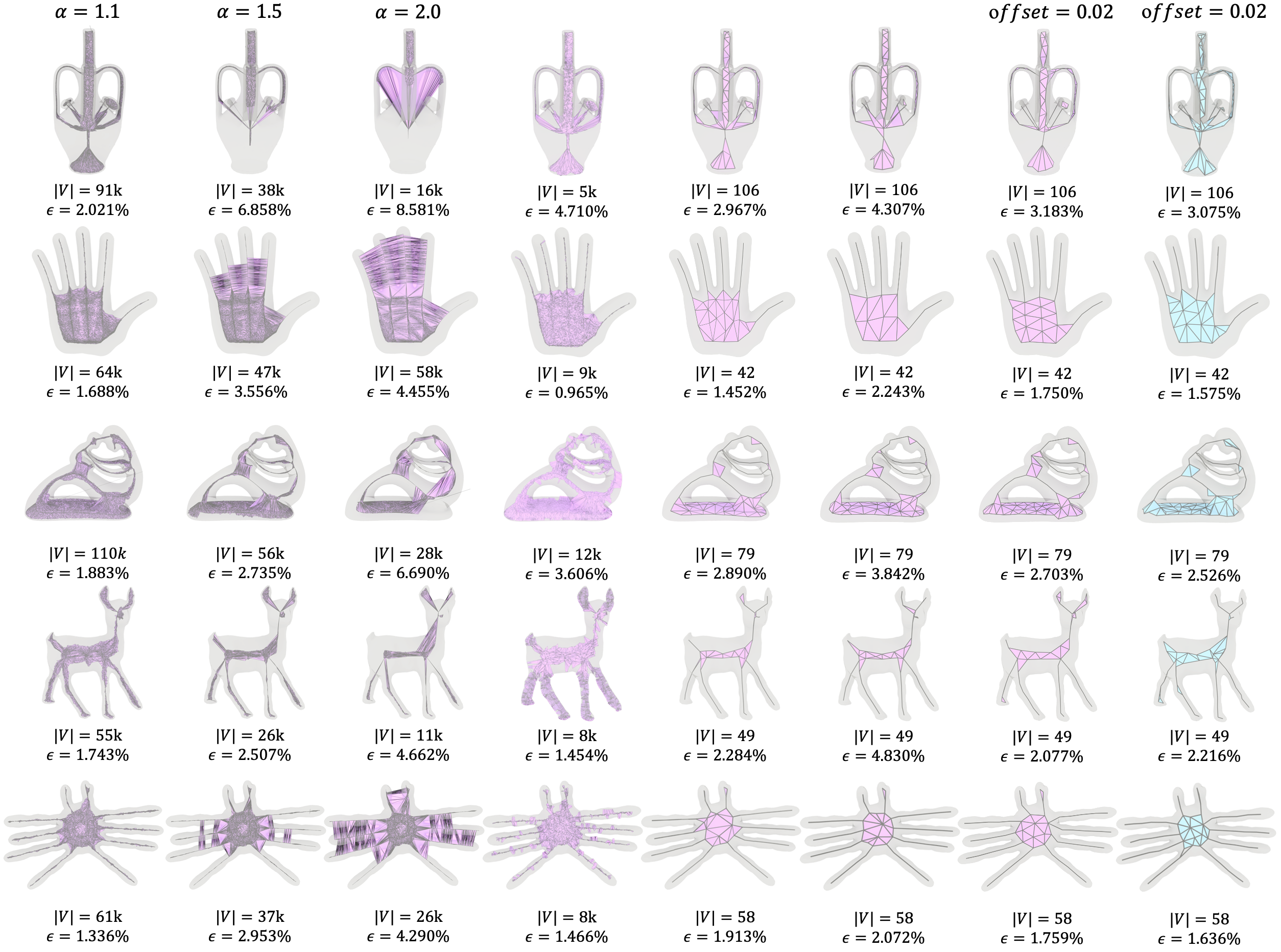} 
\put(6.2,-2){(a)}
\put(18.7,-2){(b)}
\put(31,-2){(c)}
\put(43.7,-2){(d)}
\put(56.2,-2){(e)}
\put(68.2,-2){(f)}
\put(80.7,-2){(g)}
\put(92.7,-2){(h)}
\end{overpic}
\vspace{0mm}
     \caption{Qualitative comparisons. (a-c) SAT. (d) MATFP. (e) QMAT. (f) QMAT+. (g) Coverage Axis. (h) Coverage Axis++.}
\label{fig:sat}
\end{figure*}
\begin{figure*}[]
\centering
\includegraphics[width=0.945\textwidth]{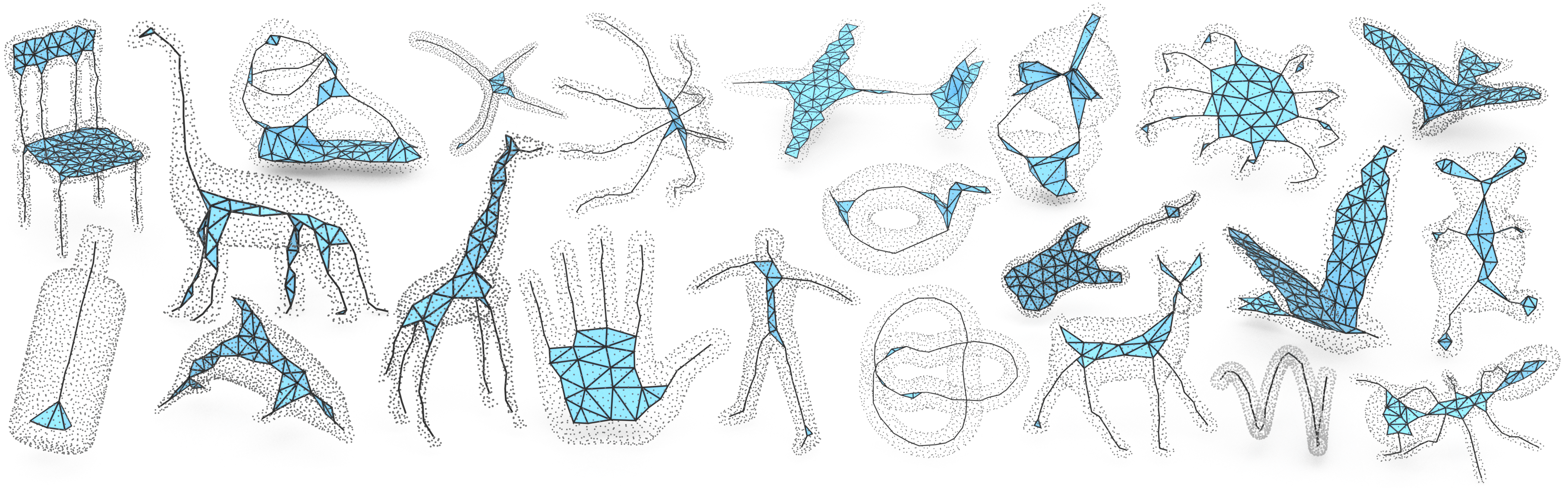} 
\vspace{-2mm}
  \caption{A gallery of skeletonization results of point clouds using Coverage Axis++.}
  \label{fig:pc_collection}
\end{figure*}

\section{Experimental Results}
\label{sec:exp_res}
We conduct extensive qualitative and quantitative evaluations of the proposed method to demonstrate its effectiveness comprehensively. All experiments are conducted on a computer with a 4.00 GHz Intel(R) Core(TM) i7-6700K CPU and 32 GB memory. The sizes of all the models are normalized to the $[0,1]$ range.  

\begin{figure*}[]
\centering
\begin{overpic}[width=0.97\linewidth]{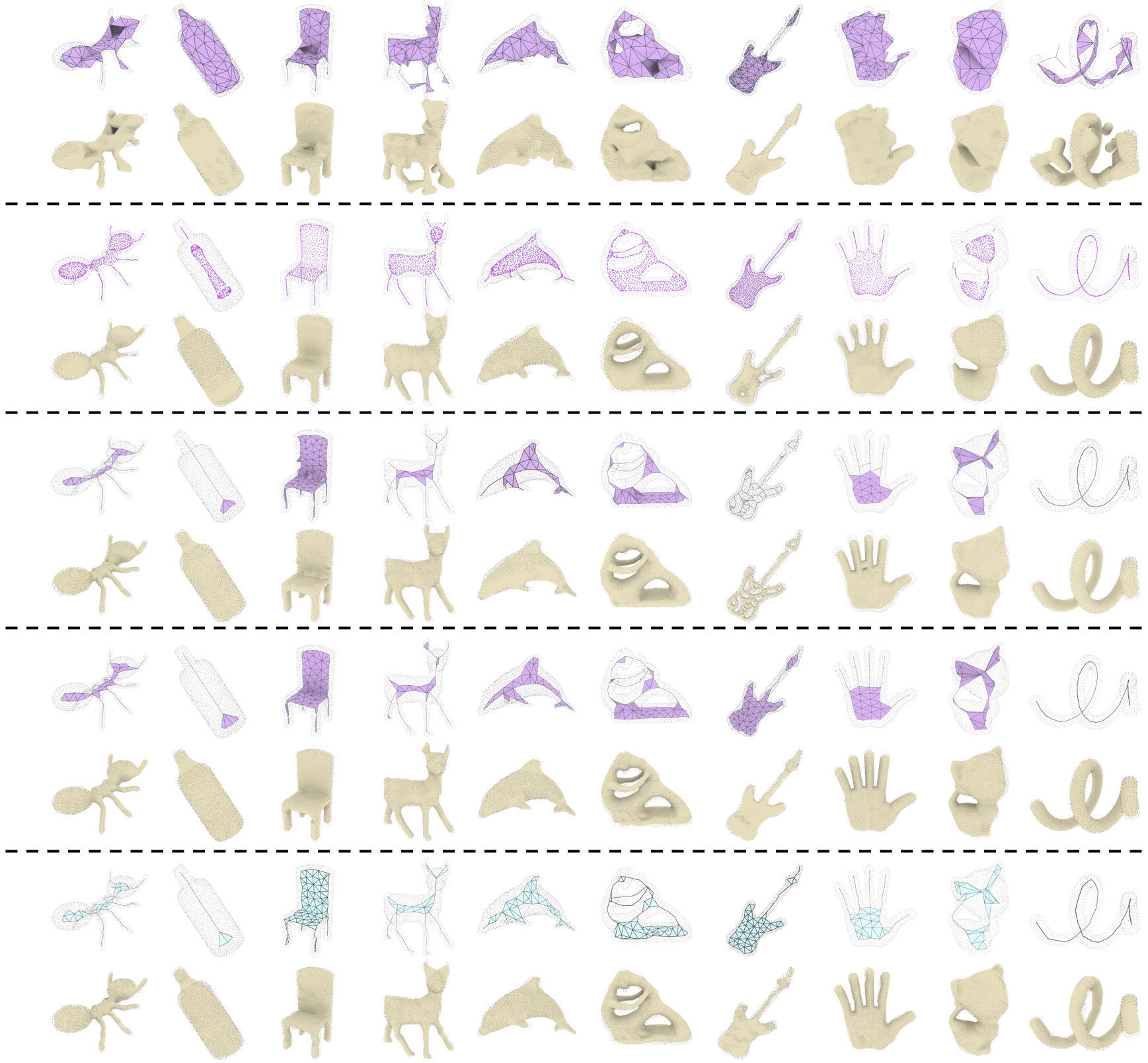} 
\put(1.5,78){\rotatebox{90}{Point2Skeleton}}
\put(1.5,64){\rotatebox{90}{DPC}}
\put(1.5,41){\rotatebox{90}{Neural Skeleton}}
\put(1.5,23){\rotatebox{90}{Coverage Axis}}
\put(1.5,2.5){\rotatebox{90}{Coverage Axis++}}
\end{overpic}
\vspace{-1mm}
    \caption{Comparison with existing shape skeletonization methods for point clouds.}
\label{fig:pc}
\vspace{-6mm}
\end{figure*}

Given the target skeletal point number, Coverage Axis++ only includes one hyperparameter: $\delta_r$ (Eq.~\ref{eq:catplus_dilation}). All the experimental results are generated using a consistent parameter setting, i.e., for all kinds of input, we always set $\delta_r = 0.02$.
In experiments where we compare our method with Coverage Axis, we set the number of selected points $|V|$ for Coverage Axis++ to be equal to the number obtained from the SCP solution for the Coverage Axis. We provide additional analyses on parameters $\delta_r$ and $|V|$ in Sec.~\ref{sec:ablation_study}.

As suggested by ~\cite{dou2022coverage}, we employ the two-sided Hausdorff distance (HD) to assess the consistency between the original and reconstructed surfaces. We use $\overrightarrow{\epsilon}$ for surface-to-reconstruction HD, $\overleftarrow{\epsilon}$ for reconstruction-to-surface HD, and $\overleftrightarrow{\epsilon}$ for two-sided HD. The reconstructed surfaces are obtained by interpolating medial balls based on skeleton connectivity, akin to~\cite{li2015q, dou2022coverage, wang2022restricted}. Errors are normalized by the diagonal length of the input model bounding box. In quantitative results, the best results for runtime and accuracy are indicated with blue and orange highlighting, respectively. A gallery of skeletonization results of mesh and point cloud inputs are shown in Fig.~\ref{fig:teaser} and Fig.~\ref{fig:pc_collection}, respectively.

\subsection{Mesh Inputs}
\subsubsection{Comparison with Coverage Axis}
As Coverage Axis~\cite{dou2022coverage} can handle both mesh and point cloud inputs, we compared the proposed method with it on both two inputs. The quantitative results are summarized in Tab.~\ref{table:QMAT}. As can be observed, for both mesh and point cloud inputs, Coverage Axis++ requires less runtime while maintaining competitive accuracy on most shapes. \ZY{We observed that in specific cases, such as the Vase model, the Coverage Axis shows a speed advantage over our approach. This advantage is primarily attributed to the efficiency of its heuristic algorithm\footnote{\ZY{Mixed-integer linear programming in SciPy or MATLAB typically employs heuristics to search for integer-feasible solutions.}} for solving the Set Cover Problem (SCP). However, such instances are infrequent due to the theoretical computational complexity associated with SCP. When evaluating the overall performance,} our method significantly outperforms the Coverage Axis in terms of running time, where ours achieves $5.362\%$ reconstruction error in $2.9$s on average, whereas the Coverage Axis achieves $6.574\%$ reconstruction error in more than $236.1$s. The same advantage is also reflected in point cloud inputs. Actually, the runtime for Coverage Axis tends to be longer for shapes containing planar structures due to the complexity of the algorithm; See discussion in Appendix~D. The running times of the Coverage Axis are even longer when using more surface samples; See Appendix.~A. A qualitative comparison is provided in Fig.~\ref{fig:pc}.

\begin{table*}
\small
\begin{center}
\caption{Comparison of runtime and shape approximation error among Q-MAT, Q-MAT+, and Coverage Axis++ using mesh inputs.}
\label{table:QMAT2}
\vspace{-2mm}
\resizebox{\textwidth}{!}{
\setlength{\tabcolsep}{2.3mm}{
\begin{tabular}{p{1.5cm}|p{0.8cm}<{\centering}|cccc|cccc|cccc}
\hline
\multirow{2}{*}{Model} & \multirow{2}{*}{\makecell[c]{$|V|$} }   &  \multicolumn{4}{c|}{Q-MAT} &\multicolumn{4}{c|}{Q-MAT+} & \multicolumn{4}{c}{Coverage Axis++}\\
  &  &  \textbf{Time (s)} & $\overrightarrow{\epsilon}$ &  $\overleftarrow{\epsilon}$ &  $\overleftrightarrow{\epsilon}$ 
    &\textbf{Time (s)} &  $\overrightarrow{\epsilon}$ &  $\overleftarrow{\epsilon}$ &  $\overleftrightarrow{\epsilon}$ 
    &\textbf{Time (s)} &  $\overrightarrow{\epsilon}$ &  $\overleftarrow{\epsilon}$ &  $\overleftrightarrow{\epsilon}$ 
    \\
    \hline
Armadillo & $82$ & $9.8$ & $2.736\%$ & $5.678\%$ &	$5.678\%$ & $13.2$ & $3.513\%$ & $5.321\%$ & $5.321\%$ & \toptime{$1.8$} & $3.615\%$ & $5.106\%$ & \topac{$5.106\%$} \\
Bear & $49$  & $1.6$ & $3.133\%$ & $3.089\%$ & $\topac{3.133\%}$ & 2.2 & $3.480\%$ & $3.550\%$ & $3.550\%$ & \toptime{$1.0$} & $4.313\%$ & $4.572\%$ & $4.572\%$ \\
Bird  &$83$  & $2.3$ & $2.361\%$	& $3.936\%$ & $3.936\%$ & 3.1 & $2.519\%$ & $4.034\%$ & $4.034\%$ &	\toptime{$2.3$} & $2.306\%$ & $2.560\%$ & \topac{$2.560\%$} \\ 
Bunny &$73$  & $16.6$ & $3.474\%$ & $4.686\%$ & $4.686\%$ & 19.7 & $3.501\%$ & $3.762\%$ & \topac{$3.762\%$} &\toptime{ $1.5$} & $4.860\%$ & $5.771\%$	& $5.771\%$ \\
Camel & $93$ & $2.8$ & $2.204\%$ & $2.468\%$ & $2.468\%$ & 4.1 & $2.047\%$ & $2.110\%$ & \topac{$2.110\%$}&\toptime{$2.5$} & $2.175\%$ & $5.556\%$ & $5.556\%$ \\
Dinosaur & $66$  & $8.1$ & $1.930\%$ & $2.444\%$ & \topac{$2.444\%$} & 9.9 & $1.942\%$ & $3.408\%$ & $3.408\%$ & \toptime{$1.4$} & $2.308\%$ & $2.672\%$ & $2.672\%$ \\
Dog	& $69$  & \toptime{$1.3$} & $2.299\%$ & $3.781\%$ & $3.781\%$& 1.7 & $2.173\%$ & $2.725\%$ & \topac{$2.725\%$} & $1.5$ & $3.353\%$ & $3.799\%$ &	$3.799\%$ \\ 
Fandisk	 & $127$ & \toptime{$2.9$} &	$4.888\%$ & $5.840\%$ &	$5.840\%$ & 3.8 & $4.911\%$ & $5.920\%$ & $5.920\%$& $4.6$ &	$3.849\%$ & $4.067\%$ & \topac{$4.067\%$} \\
Hand & $47$  & $21.9$ & $1.441\%$ & $1.087\%$ &\topac{$1.441\%$} & 25.5 & $2.173\%$ & $2.208\%$ & $2.208\%$& \toptime{$1.1$} & $1.974\%$ & $1.975\%$ & $1.975\%$ \\
Horse  & $81$   & $22.9$ & $1.973\%$ & $3.683\%$ & $3.683\%$& 24.4 & $2.362\%$ & $3.853\%$ &  $3.853\%$ & \toptime{$1.9$} & $2.382\%$ & $2.658\%$ & \topac{$2.658\%$} \\
Human & 46 &$1.6$ & $2.006\%$ & $2.155\%$ &$2.155\%$  & 2.1 & $1.545\%$ & $2.217\%$ & $2.217\%$ &	\toptime{$0.9$} & $2.062\%$ &	$2.105\%$ & \topac{$2.105\%$} \\
Neptune & $53$ & $5.0$ & $2.379\%$ & $11.372\%$ & $11.372\%$& 7.2 & $4.229\%$ & $7.146\%$ & \topac{$7.146\%$} &\toptime{$1.2$} & $3.227\%$ & $9.989\%$ & $9.989\%$ \\
Seahorse & $60$  &$7.6$ & $1.908\%$ & $4.731\%$ & $4.731\%$ & 8.9 & $1.806\%$ & $4.557\%$ & \topac{$4.557\%$}& \toptime{$1.4$} & $2.296\%$ & $5.168\%$ & $5.168\%$ \\
Vase & $118$  & $6.2$ & $2.826\%$ & $4.839\%$ & $4.839\%$ & 8.0 & $3.038\%$ & $3.893\%$ & $3.893\%$ & \toptime{$3.3$} & $3.575\%$ & $3.875\%$ & \topac{$3.875\%$} \\
\hline
\end{tabular}}
}
\end{center}
\vspace{-2mm}
\end{table*}

\subsubsection{Comparison with MATFP} 
We further compare our method with the recent work MATFP~\cite{wang2022computing}. MATFP aims to compute MAT with feature preservation, and it only takes watertight mesh input.  Besides, similar to the Coverage Axis, the user cannot explicitly specify the final number of skeletal points. As shown in Fig.~\ref{fig:sat}, we find MATFP typically yields results with redundant points. In comparison, Coverage Axis++ produces more compact results while using less computation time (13.11s v.s. ours 2.61s). Since MATFP is a multi-stage approach, we detail the running time of each stage in Appendix~B.

\subsubsection{Comparison with Q-MAT and Q-MAT+} 
We compare our method with two representative methods for MAT computation given the mesh inputs: Q-MAT~\cite{li2015q} and its variant Q-MAT+~\cite{pan2019q}. Both two methods take watertight meshes as inputs and cannot handle point clouds or polygon soups. Tab.~\ref{table:QMAT2} shows that Coverage Axis++ performs comparably or superior to Q-MAT and Q-MAT+ in terms of reconstruction accuracy and running time. Note that both Q-MAT and Q-MAT+ require a watertight mesh for Voronoi initialization, while our method supports \textit{randomly} generated candidate points inside the volume.

\subsubsection{Comparison with SAT} 
SAT~\cite{miklos2010discrete} encounters a major drawback as it tends to favor a dense representation with numerous vertices, hindering the generation of a simple and compact skeleton of the given shape. Note SAT only excels specifically with watertight surface meshes. We test with different scaling factors, $\alpha = 1.1$, $1.5$, and $2.0$, revealing that with a smaller $\alpha$, SAT achieves high accuracy but results in a representation with a large number of vertices (Fig.\ref{fig:sat}(a-c)). Using a larger $\alpha$ for higher abstraction compromises shape structure and introduces significant approximation errors. As a comparison, Coverage Axis++ achieves a balance between fidelity, efficiency, and compression abilities.

\subsection{Point Cloud Inputs}
\subsubsection{Comparison with Neural Skeleton} 
We compare Coverage Axis++ with Neural Skeleton~\cite{clemot2023neural} given the point cloud inputs. Neural Skeleton first trains a 
SDF w.r.t. the input point cloud, then samples uniform surface points and skeletal points from the implicit field. Finally, the skeleton mesh is recovered using a cover-set formulation~\cite{dou2022coverage} solved as a Mixed-Integer Linear Program. The implicit field improves the robustness of the Neural Skeleton in handling challenging inputs. As shown in Tab.~\ref{table:pc}, our method produces results with higher accuracy. Moreover, the average running time of Coverage Axis++ and NeuralSkeleton is \textit{1.63}s v.s. \textit{59.27}s. The detailed runtime statistics of Neural Skeleton are given in Appendix~B. Note that the Neural Skeleton does not support specifications for the number of skeletal points.

\subsubsection{Comparison with Point2Skeleton}
Compared with Point2Skeleton~\cite{lin2021point2skeleton}, a learning-based skeletonization method for point clouds, our approach exhibits better performance. 
As shown in  Fig.~\ref{fig:pc} and Tab.~\ref{table:pc}, Point2Skeleton, trained on a large dataset~ShapeNet~\cite{chang2015shapenet}, falls short on shapes unseen during training and struggles with higher topology complexity. However, our method demonstrates robustness, generating high-quality skeletal representations with accurate geometries and faithful structures across diverse shapes.

\subsubsection{Comparison with Deep Point Consolidation}  
As shown in Fig.~\ref{fig:pc} and Tab.~\ref{table:pc}, Deep Point Consolidation (DPC)~\cite{wu2015deep} can only produce unstructured inner points without connections and the reconstruction results exhibit large errors.  Even worse, the reconstructed topologies by DPC are usually inconsistent with the original input. In contrast, our reconstruction results reach better approximation accuracy with respect to the original geometry using fewer skeletal points. 

\begin{table}[t]
\small
\begin{center}
\caption{Comparison on shape approximation errors among Point2Skeleton (P2S), Deep Point Consolidation (DPC), Neural Skeleton, Coverage Axis and Coverage Axis++.}
\label{table:pc}
\vspace{-2mm}
\resizebox{0.485\textwidth}{!}{
\begin{tabular}{p{1.18cm}|cc|cc|cc|cc|cc}
\hline
\multirow{2}{*}{Model}  &  \multicolumn{2}{c|}{P2S} &\multicolumn{2}{c|}{DPC} & \multicolumn{2}{c|}{Neural Skeleton} & \multicolumn{2}{c|}{Coverage Axis}  & \multicolumn{2}{c}{Coverage Axis++}\\
 & $|V|$ &  $\overleftrightarrow{\epsilon}$ 
  & $|V|$ &  $\overleftrightarrow{\epsilon}$ & $|V|$ &   $\overleftrightarrow{\epsilon}$ & $|V|$ &  $\overleftrightarrow{\epsilon}$ & $|V|$ &  $\overleftrightarrow{\epsilon}$\\
    \hline
Ant-2  & $100$ & $ 16.412\%$ & $ 1194$ & $ 8.863 \%$& 54 & $3.993\%$ & $58$ & $ 2.350\%$  & $58$ & $ \topac{2.113\%}$ \\ 
Bottle  & $100$ & $2.955\%$ & $1194$ &$ 2.752\%$ & 13 & \topac{$2.390\%$} & $14$ & $  2.956\%$ & $14$ & $2.599\%$\\
Chair-2 & $100$ & $6.552\%$ & $1194$ & $ 4.807\%$ & 134 & $3.651\%$ & $ 89$ & $ \topac{2.890\%}$ & $89$ & $3.012\%$\\
Dog  & $100$ & $6.047\%$ & $ 1194$ & $ 5.237 \%$ & 42 & $3.074\%$ & $49$ & $ \topac{2.174\%}$ & $49$ & $2.306\%$\\
Dolphin &$100$ & $5.925\%$ & $ 1194$ & $ 7.916 \%$& 36 &$3.269\%$ & $ 49$ & $ 1.971\%$ & $49$ & $\topac{1.876\%}$\\
Fertility & $100$ & $8.162\%$ & $ 1194 $ & $ 4.226 \%$ & 76 &$5.909\%$ & $ 79$ & $ 3.428\%$ & $79$ & $\topac{3.102\%}$\\
Guitar & $100$ & $ 2.052\%$ & $1194$ & $3.231\%$ & 65 & $2.872\%$ & $60$ & $ \topac{2.032\%}$ & $60$ & $2.344\%$\\
Hand &  $100$ & $9.672\%$ & $ 1194$ & $ 4.110 \%$ & 39 &$4.819\%$ & $ 44$ & $ 3.441\%$ & $44$ & $\topac{2.328\%}$\\
Kitten & $100$ & $8.724\%$ & $ 1194$ & $ 5.332 \%$ & 41 &$3.883\%$ & $ 50$ & $3.450\%$ & $50$ & $\topac{2.962\%}$\\
Snake & $100$ & $15.021\%$  & $ 1194$ & $  1.736 \%$ & 36 & $1.419\%$ & $ 40$ & $ \topac{1.309\%}$ & $40$ & $1.402\%$\\
\hline
Average  & - & $8.003\%$ & - & $  4.815 \%$ &  - & $3.528\%$ & - & $ 2.515\%$ & -& $\topac{2.404\%}$\\
\hline
\end{tabular}}
\end{center}
\end{table}

\section{More Discussions}
\label{sec:ablation_study}
\begin{figure}[t]
    \centering
  \begin{overpic}
[width=\linewidth]{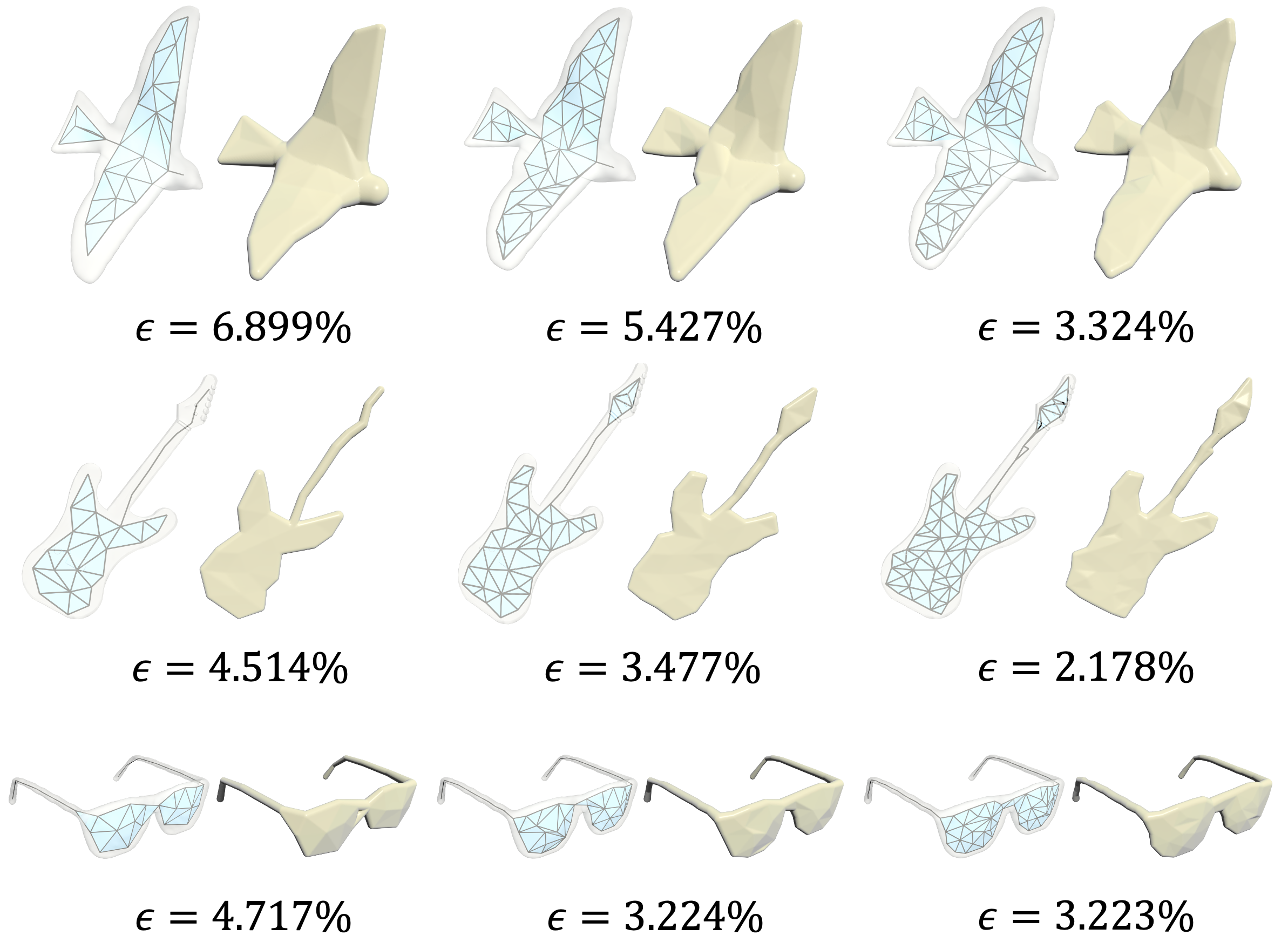}
\put(12,-3){$|V|=30$}
\put(44,-3){$|V|=50$}
\put(77.5,-3){$|V|=70$}
\end{overpic}
 \vspace{-2mm}
    \caption{Influence of target numbers $|V|$. We test our model with the target numbers being $30$, $50$ and $70$.}
    \label{fig:skeleton_num}
\end{figure}
\subsection{Skeletal Point Number}
\label{sec:exp_num_skeleton}

In Sec.~\ref{subsec:more_features}, we highlight Coverage Axis++'s practical benefit of flexible specification for the target number of skeletal points. Fig.~\ref{fig:skeleton_num} presents our experiments with different target skeletal numbers using three models and three choices of $|V|$. Larger values of $|V|$ result in more accurate skeletal representations, as observed in the experiments where reconstruction error consistently decreases with increasing $|V|$. However, choosing excessively large $|V|$ contradicts our objective of generating a simple and compact skeleton for shape abstraction. Our results show that specifying a smaller number of skeletal points (e.g., $30$ or $50$) achieves simpler shape abstraction with only a slight increase in reconstruction error. The flexibility allows users to tailor the method for either a simple or more accurate model, depending on the requirement for shape abstraction.

\subsection{Dilation Factor}
\label{sec:dilation_factor}
\begin{figure}
    \centering  
\begin{overpic}
[width=\linewidth]{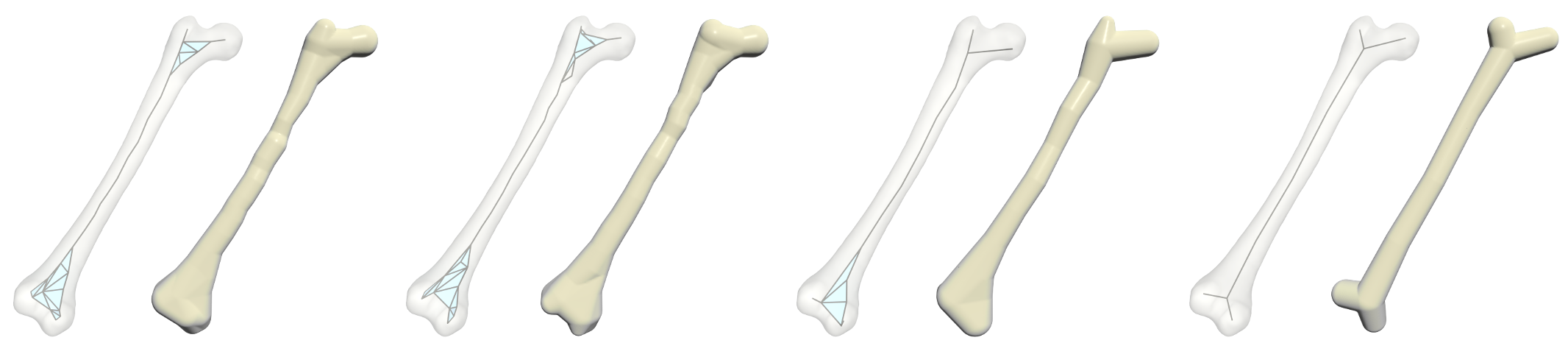}
\put(5.5,-3.7){$|V|=30$}
\put(30,-3.7){$|V|=30$}
\put(55,-3.7){$|V|=14$}
\put(80,-3.7){$|V|=7$}
\put(5,-8){$\delta_r=0.01$}
\put(29,-8){$\delta_r=0.02$}
\put(54,-8){$\delta_r=0.05$}
\put(79,-8){$\delta_r=0.1$}
\put(4,-12){$\epsilon=3.211\%$}
\put(28,-12){$\epsilon=2.101\%$}
\put(53,-12){$\epsilon=3.236\%$}
\put(78,-12){$\epsilon=5.304\%$}
\end{overpic}
 \vspace{5mm}
    \caption{Skeletonization results using different dilation offsets.}
    \label{fig:delta}
\end{figure}

Next, we perform an ablation study on the dilation factor $\delta_r$ using the Femur model as an example. As shown in Fig.~\ref{fig:delta}, increasing $\delta_r$ allows balls to cover more surface samples, simplifying the skeletal representation. We always set $|V|=30$ in Algorithm \ref{heuristic}, but when $\delta_r$ is $0.05$ or larger, the algorithm terminates with fewer than 30 selected points as dilated balls already cover all surface samples. A larger $\delta_r$ simplifies the representation but sacrifices geometric details, compromising shape structure preservation. Conversely, an excessively small $\delta_r$ contradicts simplicity and compactness. We maintain consistency by setting $\delta_r$ to $0.02$ in all experiments, balancing geometric details and skeletal compactness.

\begin{figure*}[]
    \vspace{12mm}
\centering
\begin{overpic}[width=0.97\linewidth]{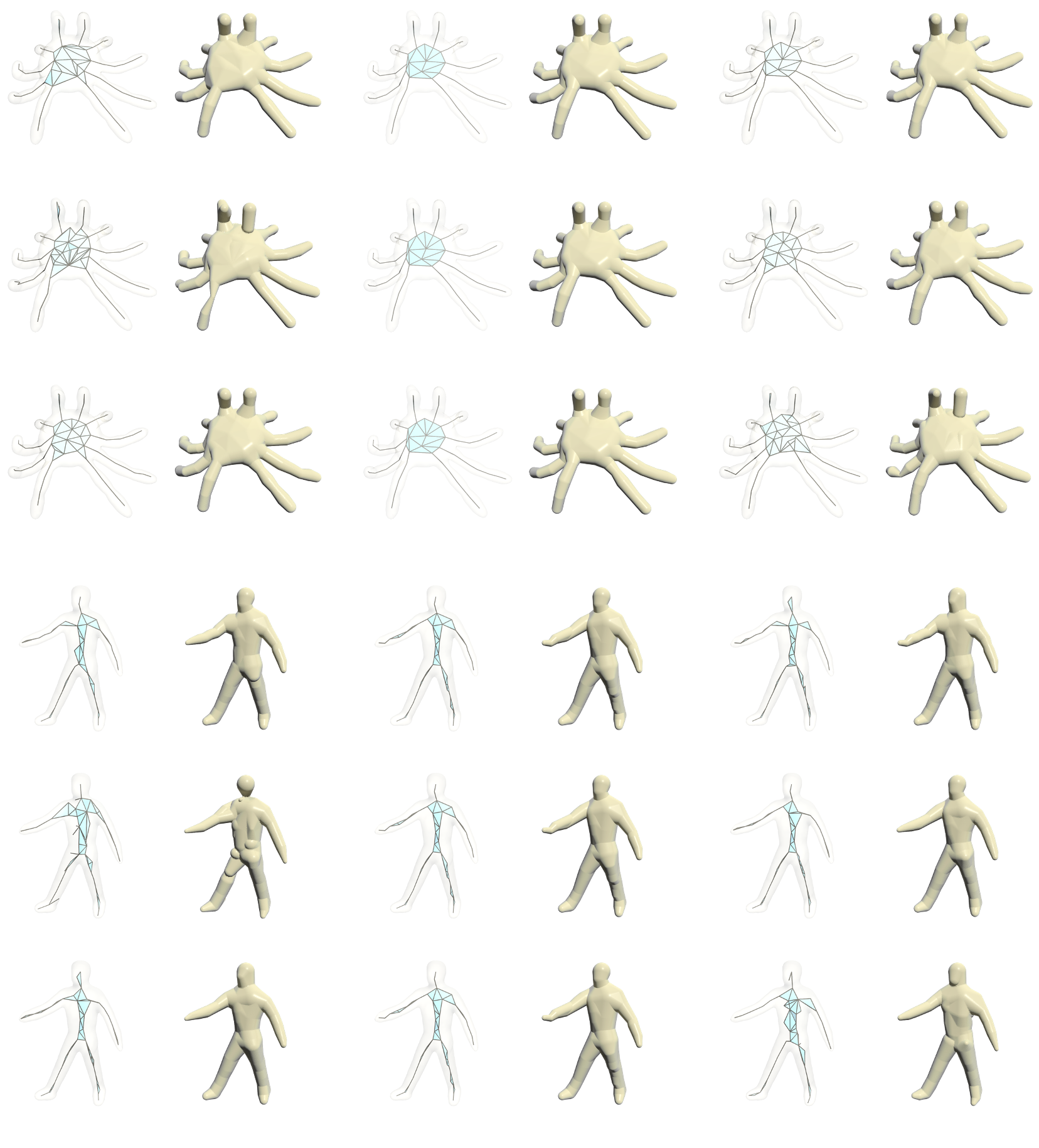} 
\put(10,86){$\epsilon=3.986\%$}
\put(9.5,84){(1.0, 1.0, \underline{5.0})}
\put(10,69.5){$\epsilon=4.457\%$}
\put(9.5,67.5){(1.0, \underline{5.0}, 1.0)}
\put(10,53){$\epsilon=3.826\%$}
\put(9.5,51){(\underline{5.0}, 1.0, 1.0)}

\put(40.5,86){$\epsilon=3.796\%$}
\put(40,84){(1.0, 1.0, \underline{1.0})}
\put(40.5,69.5){$\epsilon=3.796\%$}
\put(40,67.5){(1.0, \underline{1.0}, 1.0)}
\put(40.5,53){$\epsilon=3.796\%$}
\put(40,51){(\underline{1.0}, 1.0, 1.0)}

\put(71.5,86){$\epsilon=3.796\%$}
\put(71,84){(1.0, 1.0, \underline{0.2})}
\put(71.5,69.5){$\epsilon=3.826\%$}
\put(71,67.5){(1.0, \underline{0.2}, 1.0)}
\put(71.5,53){$\epsilon=3.821\%$}
\put(71,51){(\underline{0.2}, 1.0, 1.0)}

\put(10,35){$\epsilon=2.512\%$}
\put(9.5,33){(1.0, 1.0, \underline{5.0})}
\put(10,18.5){$\epsilon=4.982\%$}
\put(9.5,16.5){(1.0, \underline{5.0}, 1.0)}
\put(10,2){$\epsilon=2.422\%$}
\put(9.5,0){(\underline{5.0}, 1.0, 1.0)}

\put(40.5,35){$\epsilon=2.392\%$}
\put(40,33){(1.0, 1.0, \underline{1.0})}
\put(40.5,18.5){$\epsilon=2.392\%$}
\put(40,16.5){(1.0, \underline{1.0}, 1.0)}
\put(40.5,2){$\epsilon=2.392\%$}
\put(40,0){(\underline{1.0}, 1.0, 1.0)}

\put(71.5,35){$\epsilon=2.429\%$}
\put(71,33){(1.0, 1.0, \underline{0.2})}
\put(71.5,18.5){$\epsilon=2.422\%$}
\put(71,16.5){(1.0, \underline{0.2}, 1.0)}
\put(71.5,2){$\epsilon=2.857\%$}
\put(71,0){(\underline{0.2}, 1.0, 1.0)}

\put(-1.5,88){\rotatebox{90}{Centrality Score}}

\put(-1.5,71){\rotatebox{90}{Uniformity Score}}

\put(-1.5, 55.5){\rotatebox{90}{Coverage Score}}

\put(-1,37.5){\rotatebox{90}{Centrality Score}}

\put(-1,20){\rotatebox{90}{Uniformity Score}}

\put(-1,4.5){\rotatebox{90}{Coverage Score}}

\end{overpic}
\vspace{1mm}
    \caption{\ZY{We conducted ablation experiments on different weights. We test our method using different weights corresponding to the centrality score, uniformity score, and coverage score, e.g., $(1.0,1.0,0.2)$ refers to $Score_i = 1.0 \cdot Cov_i + 1.0 \cdot Unif_i + 0.2 \cdot Centr_i$. We visualize the skeletal results and evaluate the reconstruction error. We test with Human and Crab models.}}
\label{fig:diff_weights_ablation}
\vspace{-4mm}
\end{figure*}

\begin{figure}
    \centering
  \begin{overpic}
[width=\linewidth]{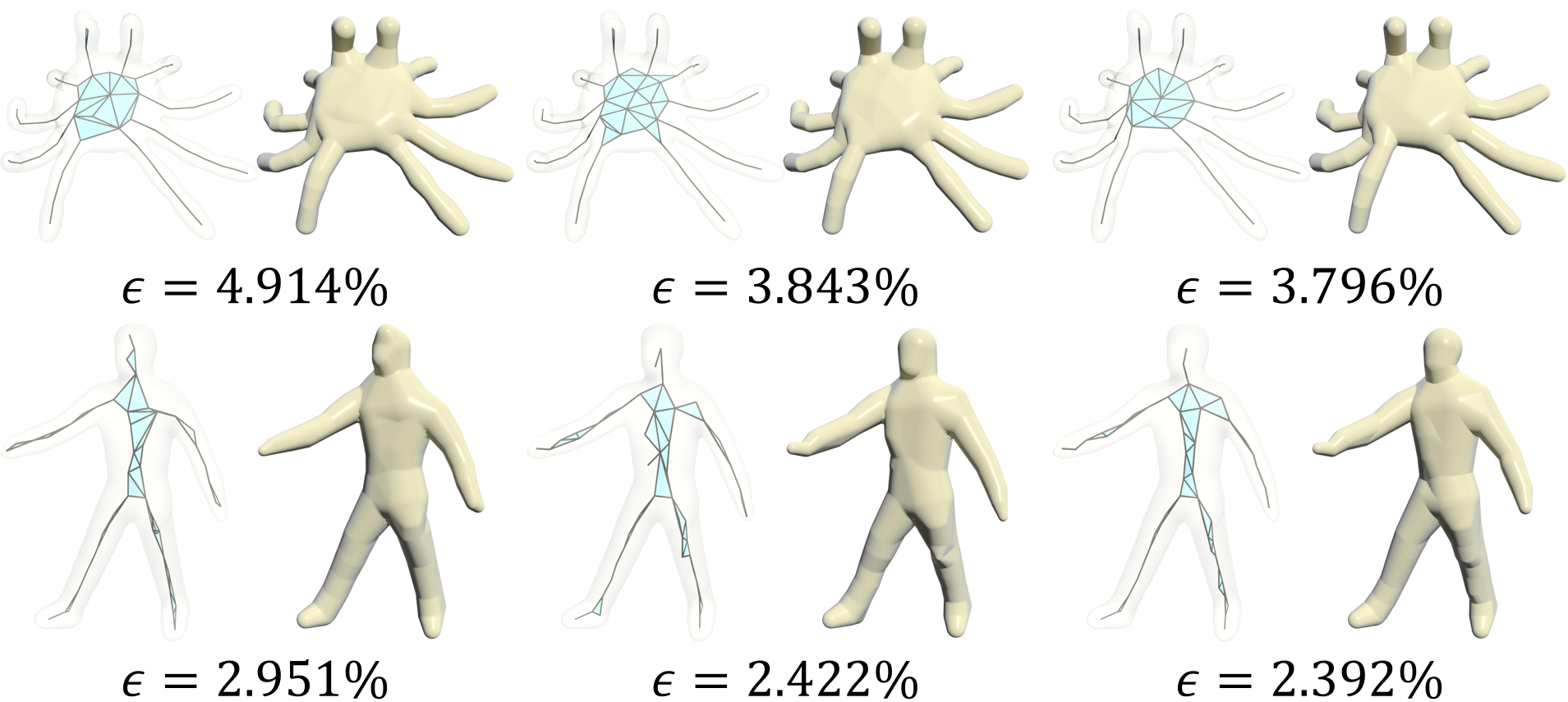}
\put(14,-5){(a)}
\put(48,-5){(b)}
\put(81,-5){(c)}
\end{overpic}
 \vspace{-1mm}
    \caption{Performance of different combinations of the three scores. (a) \textit{coverage} score. (b) \textit{coverage} + \textit{uniformity} score. (c) \textit{coverage} + \textit{uniformity} + \textit{centrality} score (Eq.~\ref{eq:score}).}
    \label{fig:score_weight}
\end{figure}

\subsection{Ablation on Scoring Terms}
\label{subsubsec:scoring}
We evaluate our method with different scoring terms defined as different combination schemes of the \textit{coverage} (Eq.~\ref{eq:cover_score}), \textit{uniformity} (Eq.~\ref{eq:uniform})and \textit{centrality} (Eq.~\ref{eq:reg}) scores on the Crab and Human models, where we set $|V|=60$ for both models. 
From Fig.~\ref{fig:score_weight}, we observe that the inclusion of the uniformity score results in skeletonizations with fewer sharp triangles, indicating a more uniform distribution of selected points. Furthermore, incorporating the centrality score enables the selected skeletal points to capture more accurate MAT information, leading to a more compact and regular skeletonization. Consequently, incorporating all three scores, as shown in Eq.~\ref{eq:score} yields improved skeletonization outcomes, i.e., compact and high-accuracy shape approximation.

\ZY{Although using equal weights for the terms in Eq.\ref{eq:score} effectively achieves compact skeletonization and high reconstruction accuracy in our main experiments without requiring special weight adjustments, we investigate the impact of different weighting schemes for each term. Concretely, we scale the weight of each term down and up by a factor of 5, while keeping the weights of the other terms unchanged, to investigate the influence of different terms. We test our algorithm using two models: Crab and Human. As shown in Fig.~\ref{fig:diff_weights_ablation}, we observe that applying a moderate uniformity weight significantly enhances the spatial distribution uniformity of the model vertices. However, using excessively large centrality score weights may deteriorate the regular structure of skeletal points, resulting in increased reconstruction error.}

\subsection{Shape Skeletonization for Triangle Soups}
\label{sec:soup_result}
\begin{figure}[t]
\vspace{-2mm}
    \centering
  \begin{overpic}
[width=\linewidth]{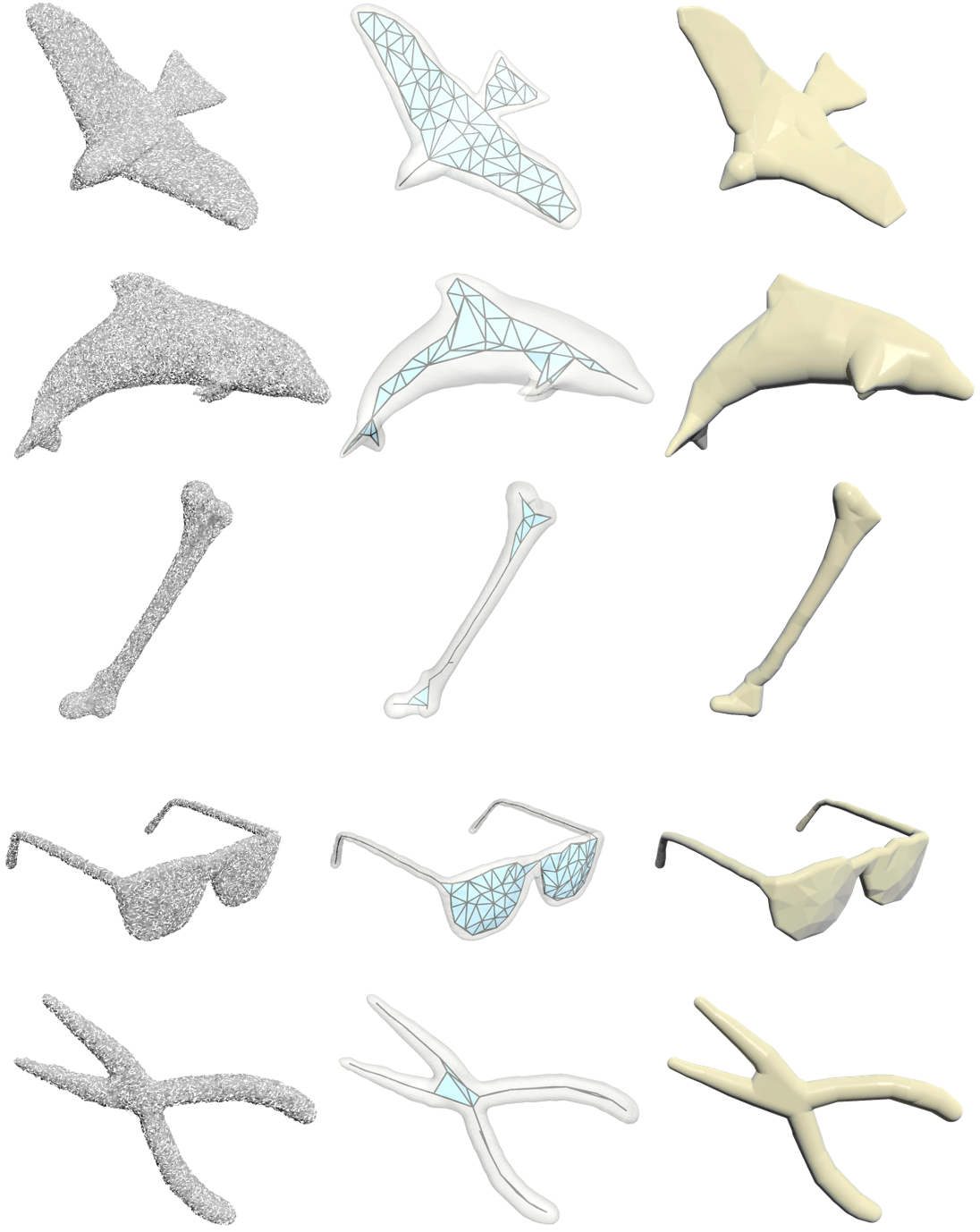}
\put(14,-3.6){(a)}
\put(41.5,-3.6){(b)}
\put(68.5,-3.6){(c)}
\end{overpic}
 \vspace{-1.5mm}
    \caption{\ZY{Shape skeletonization results for triangle soup. (a) Input triangle soup. (b) Skeletonization result. (c) Shape reconstruction.}}
    \label{fig:soup}
\end{figure}

\ZY{We evaluate our method for shape skeletonization using triangle soup inputs. For the experiment, we apply a 0.25\% perturbation relative to the bounding box to each vertex of the triangles. The skeletonization process is identical to that used for handling point cloud inputs. In this experiment, we sample $1500$ surface points from the soup and use $10000$ randomly generated candidate points inside the volume, as we described in Sec.~\ref{sec:point_generation_sec}. As shown in Fig.~\ref{fig:soup}, our method yields high-quality skeletonization and shape approximation results with the triangle soup inputs.}

\subsection{Running Time Analysis}
\label{sec:running_time_ab}
\begin{figure}
  \centering
  \small
  \subfigure[Running time w.r.t. target skeletal point numbers $|V|$ with $|S|=2000, |P|= 10000$. ]{
    \includegraphics[width=0.9\linewidth]{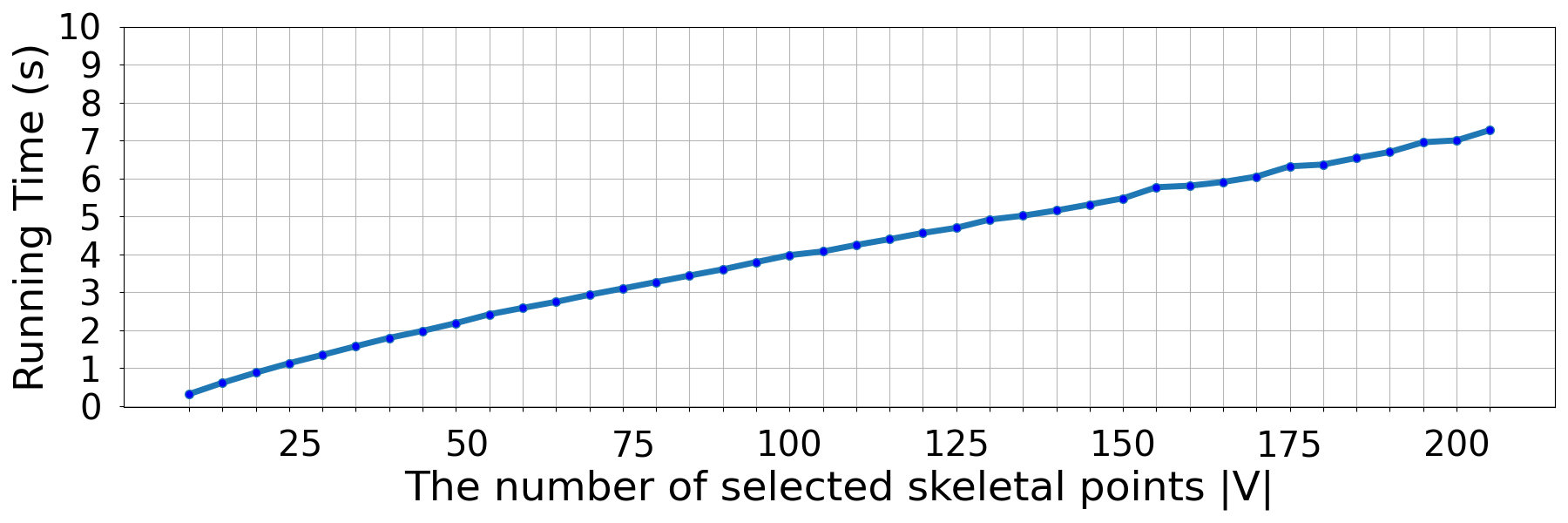}
  }
  \subfigure[Running time w.r.t. surface sample numbers $|S|$ with $|V|=100, |P|= 10000$. ]{
    \includegraphics[width=0.9\linewidth]{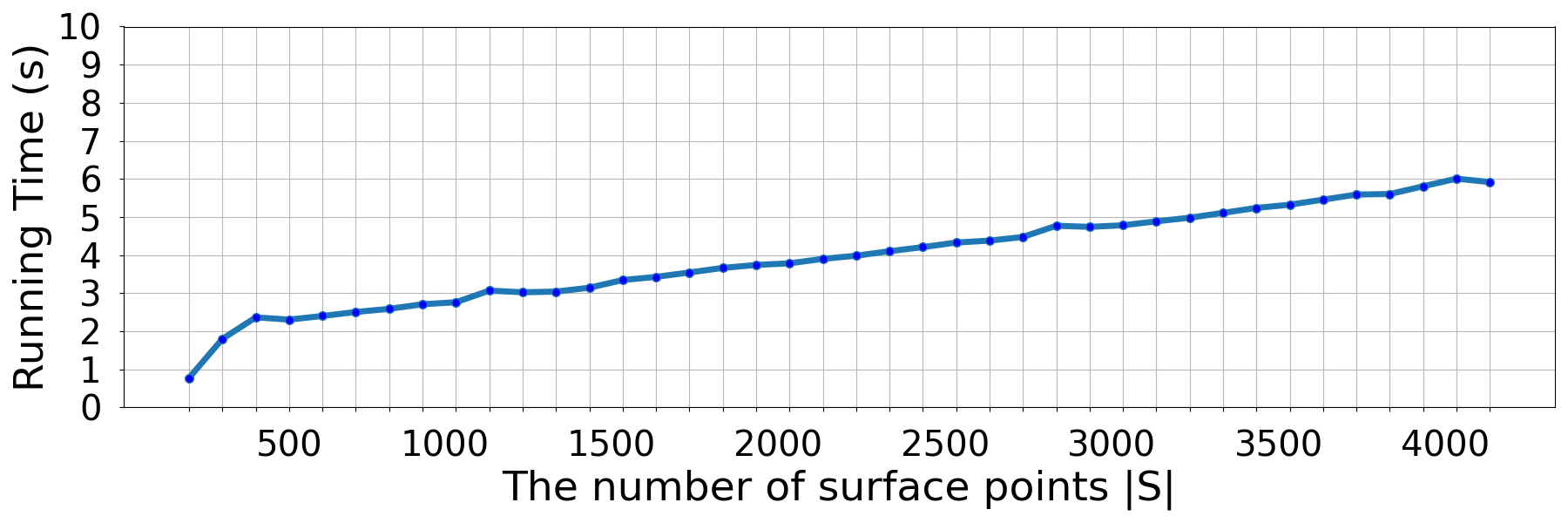}
  }
  \subfigure[Running time w.r.t. candidate skeletal point numbers $|P|$ with $|V|=100, |S|=2000$. ]{
    \includegraphics[width=0.9\linewidth]{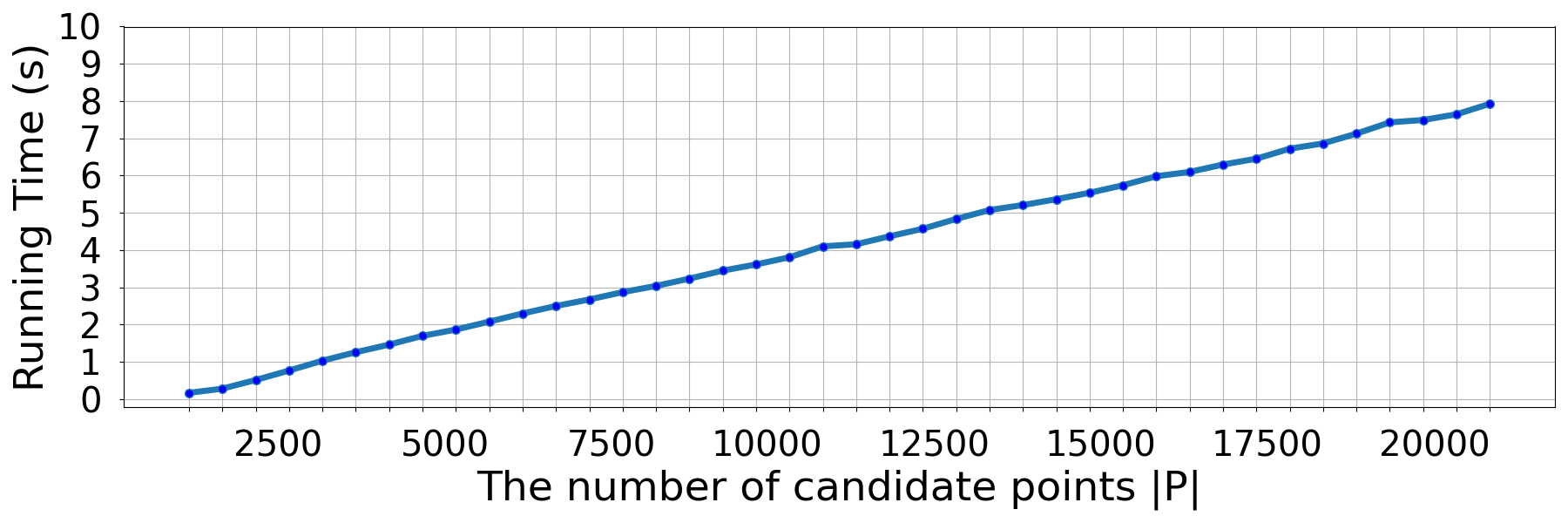}
  }
  \vspace{-3mm}
  \caption{Runtime performance w.r.t. different parameter settings.}
  \label{fig:time_ablation}
\end{figure}

The overall complexity of Algorithm \ref{heuristic} is $O(|S|\cdot|P|\cdot |V|)$. Next, we validate the practical running time complexity of our method using different set sizes $|P|$, $|S|$, and $|V|$. As shown in Fig.~\ref{fig:time_ablation}, Coverage Axis++ has polynomial complexity relationships among various variables, maintaining overall low computational costs. 

%% file: Main/5-conclusion.tex
\subsection{Limitations and Future Works}
\label{supp_sec:limitation}
Coverage Axis++ approximates the whole shape by an abstraction of local geometry with the union of the local geometries forming the entire shape. One limitation of Coverage Axis++ is that its whole process is conducted without a hard constraint forcing the consistency between the selected medial balls and the shape surface (See coverage rate statistics in Appendix.~C), like that in Coverage Axis~\cite{dou2022coverage}. \ZY{Although local approximation for shape skeletonization} has been evidenced by some previous works~\cite{lin2021point2skeleton, ge2023point2mm, amenta2001power}, we acknowledge that the inclusion of a global constraint imposes a much stricter guarantee for the shape correspondence. 
\begin{wrapfigure}{r}{2.5cm}
\vspace{-4mm}
  \hspace*{-5mm}
  \centerline{
  \includegraphics[width=31mm]{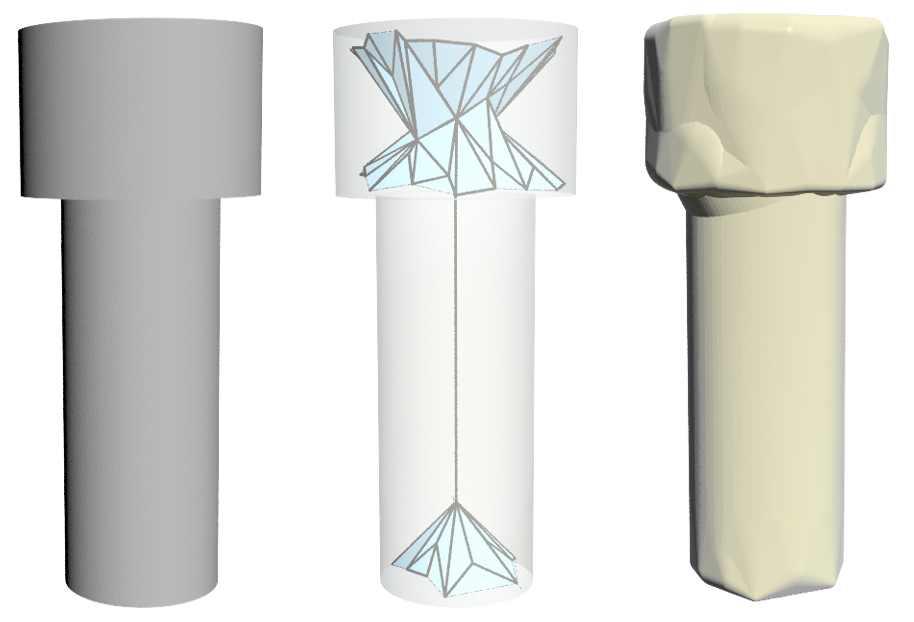}}
  \vspace*{-4mm}
\end{wrapfigure}
\ZY{Since the goal of this paper is to compute an overall compact skeletonization result, our method may encounter challenges when handling shapes like CAD models that have extremely thin structures or sharp on-surface features~(see the inset figure). In the future, introducing additional terms to preserve sharp features during shape skeletonization would be promising. For example, an algorithm can be developed to build up the connection of sharp features and the medial axis, ensuring the skeletonization result retains the sharp features on the surface.} \ZY{Our method can generate a compact representation in terms of shape coverage using medial balls as primitives. We admit that while minimizing the redundancy of skeletal points is highly desirable, there is still potential to further reduce the number of skeletal points. For example, planar components such as the body of a guitar, the seat of a chair, and the base of a fertility statue are currently oversampled because the coverage capability is limited by the medial balls that are isotropic, as shown in Fig.~\ref{fig:teaser}. Future work could explore the use of different types of bounding volume primitives, such as slab meshes~\cite{li2015q}, ellipsoids~\cite{lu2007variational}, or medial skeletal diagrams~\cite{guo2023medial} instead of medial balls, to achieve a more compact representation with fewer skeletal points, while maintaining the coverage optimization goal.}

\ZY{Next, we discuss more potential applications and future works. In this paper, we present an effective and efficient method for skeletal point selection and MAT computation.
The crucial geometrical and structural information encoded in the MAT could be helpful in various downstream applications. For instance, in the task of shape segmentation~\cite{lin2020seg}, the MAT can be leveraged for identifying various types of junctions between different parts of a 3D shape using information like manifold branch and dimensional change of the MAT without relying on 3D ground truth supervision for segmentation. In the future, we believe that a part-level shape prior derived from the MAT through self-supervised learning could benefit part-aware shape generation and reconstruction~\cite{liu2024part123}. Meanwhile, the discriminative feature represented by the MAT could also facilitate the 3D shape recognition as shown by~\cite{hu2019mat}. MAT has demonstrated its effectiveness in deep, topology-aware generation for 3D shapes~\cite{petrov2024gem3d} with the neural skeleton-based representation encoding information on both shape topology and geometry. It has also been evidenced that the topological and geometrical information of the medial mesh can be useful in remeshing and volumetric subdivision, e.g., achieving all-hexahedral mesh generation of the given shape, making the method more general than previous methods based on curve-skeleton~\cite{zhang2024medial}.

Even the point selection part of our method could serve as an effective point sampling strategy for compact shape encoding in wide applications. For example, the point selection method~\cite{dou2022coverage} has been shown to be effective in enhancing computational and memory efficiency for 3D shape representation, e.g., TetSphere Splatting~\cite{guo2024tetsphere}. More potential applications of our point selection method can be those point-based methods. Specifically, one may achieve compact 2D/3D Gaussian Splatting~\cite{ huang20242d,kerbl20233d}, also known as Gaussian pruning, given a properly defined coverage score during the training. Actually, our point selection algorithm can serve as a general sampling approach applicable to various tasks. As long as an original input representation can be expressed as \textit{candidate points} and an associated \textit{covering matrix} are properly defined, thanks to the SCP, our method could yield a more \textit{compact representation} for the given \textit{redundant input}.}

\section{Conclusion}
In this paper, we present Coverage Axis++, an efficient shape skeletonization method. Unlike existing methods that depend on input watertightness or suffer from high computational costs, Coverage Axis++ addresses these challenges by employing a heuristic algorithm considering shape coverage, uniformity, and centrality for selecting skeletal points, which ensures an efficient approximation of the MAT while significantly saving the computational cost. Compared with existing methods, our simple yet effective approach produces a compact representation of MAT, offers versatility in skeletonizing various shape representations, allows customizable specification of skeletal points, and achieves highly efficient computation with improved reconstruction accuracy. Extensive experiments across diverse 3D shapes demonstrate the efficiency and effectiveness of Coverage Axis++. 

%% file: Main/6-appendix.tex
\appendix
\renewcommand\thefigure{\Alph{section}\arabic{figure}}    
\renewcommand\thetable{\Alph{section}\arabic{table}}
\setcounter{figure}{0}
\setcounter{table}{0}

This supplementary material covers: more experimental results (Sec.~\ref{supp:more_surface}), more detailed running time statistics of existing methods (Sec.~\ref{supp_sec:runtime_more}), coverage rate of Coverage Axis++~(Sec.~\ref{supp_sec:coverage_rate}), computation complexity analysis on Coverage Axis~(Sec.~\ref{supp:cat_slow}), and an additional discussion on the skeletal point connectivity of MAT (Sec.~\ref{supp_sec:why_mesh}).

\section{More Experimental Results}
\label{supp:more_surface}
As discussed in Sec.~4.7.1, the proposed method has polynomial complexity with respect to the number of surface samples, denoted as $|S|$.
In contrast, the Coverage Axis algorithm showcases exponential complexity in $|S|$. This is evidenced by Tab.~1 of the main paper, where we report the running times of our method and Coverage Axis for various models when $|S|$ is set to $1500$. Furthermore, it can be expected that when increasing the number of surface samples, the performance gap between our method and the Coverage Axis will widen further. We conducted additional experiments by increasing the number of surface samples to $5000$ and comparing the performance of our method with the Coverage Axis across different shapes. The results are summarized in Tab.~\ref{table:more_surface}.

Note that the proportion of timeout models (i.e., models with a running time exceeding $1000$s) of Coverage Axis increases from $21.2\%$ to $50\%$ as $|S|$ increases from 1500 to 5000. Furthermore, comparing the results obtained with $|S|$ set as $1500$ and $5000$, we find that Coverage Axis++ exhibits a more significant advantage over Coverage Axis in terms of running time.  On average, our method achieves a reconstruction error of $4.429\%$ in $9.7$s, whereas Coverage Axis achieves a reconstruction error of $5.448\%$ in over $516.3$s.

\section{More Detailed Running Time Statistics.}
\label{supp_sec:runtime_more}
We provide more statistics of the running time of Neural Skeleton~\cite{clemot2023neural}, MATFP~\cite{wang2022computing}. All evaluations are performed on the same machine to ensure a fair comparison.

\begin{table}[!h]
\small
\begin{center}
\caption{Detailed running time statistics of MATFP~\cite{wang2022computing} using mesh inputs.}
\label{supp_tab:more_time_matfp}
\vspace{-2mm}
\resizebox{0.485\textwidth}{!}{
\begin{tabular}{p{1.18cm}|cccc|c}
\hline
{Model}  &  \makecell[c]{Stage1 (s)}& 
\makecell[c]{Stage2 (s)} &
\makecell[c]{Stage3 (s)}&
\makecell[c]{Stage4 (s)}&
Total Time (s)\\
    \hline
Vase & 1.62 & 3.30 & 0.66 & 2.69 & 8.27 \\
Hand & 4.15 &5.42& 1.57& 3.54& 14.68\\
Dog &  0.59 & 0.92& 1.51& 7.53& 10.55 \\
Fertility &  2.32& 4.12& 2.19 & 8.33 & 16.96 \\
Crab & 2.49 & 4.90& 1.48& 6.25& 15.12 \\
\hline
Average &  2.23 & 3.73 & 1.48 & 5.67 & 13.11 \\
\hline
\end{tabular}}
\end{center}
\vspace{-2mm}
\end{table}

\begin{table}[!h]
\small
\begin{center}
\caption{Detailed running time statistics of Neural Skeleton~\cite{clemot2023neural} using point cloud inputs.}
\label{supp_tab:more_time}
\vspace{-2mm}
\resizebox{0.485\textwidth}{!}{
\begin{tabular}{p{1.18cm}|cc|c}
\hline
{Model}  &  \makecell[c]{Network Optimization\\Time (s)}& \makecell[c]{Coverage Skeleton\\ Time (s)} & Total Time (s)\\
    \hline
Ant-2 & 9.95 & 10.53 & 20.48 \\

Bottle &  9.88 & 11.61 & 21.49 \\
Chair-2& 9.86 & 121.52 & 131.38 \\ 
Dog & 9.96 & 15.42 & 25.38 \\
Dolphin &  10.07 & 24.88 & 34.95 \\
Fertility &  9.92 & 108.15 & 118.07 \\
Guitar &  9.86 & 107.32 & 117.18 \\ 
Hand-1 &  10.01 & 21.36 & 31.37 \\ 
Kitten &  9.89 & 46.88 & 56.77 \\ 
Snake &  9.85 & 25.74 & 35.59 \\
\hline
Average &  9.92 & 49.39 & 59.27 \\
\hline
\end{tabular}}
\end{center}
\vspace{-2mm}
\end{table}

\section{Coverage Rate}
\label{supp_sec:coverage_rate}
In the main paper, we conducted extensive experiments for Coverage Axis++ on various models to demonstrate that our method is capable of extracting compact underlying shape representation of MAT while achieving low reconstruction error, as summarized in Tab.~1. Since the selected skeletal points form a good representation of the medial axis, the corresponding dilated balls should cover a significant portion of the surface samples to build the correspondence between the medial balls and the shape volume. In other words, the coverage rate of our method should be high. The idea has been verified by Coverage Axis, as it directly requires a $100\%$ coverage rate by imposing a hard constraint during the optimization. We record the coverage rates of each model for Coverage Axis++, which are summarized in Tab.~\ref{table:coverage_rate}. The result shows that our method consistently achieves very high coverage rates, with an average of $96.134\%$.

\begin{table}[!h]
\small
\begin{center}
\caption{Coverage rate of Coverage Axis++. The inputs are meshes. $|V|$: \textit{The number of skeletal points.} $\overleftrightarrow{\epsilon}$: \textit{Two-sided HD between original surface and reconstruction.}}
\label{table:coverage_rate}
\vspace{-2mm}
\setlength{\tabcolsep}{5mm}{
\begin{tabular}{p{1.3cm}|p{0.8cm}<{\centering}|c|c}
\hline
Model & $|V|$ & $\overleftrightarrow{\epsilon}$ & Coverage rate \\

\hline
    Bear-1 & $68$ & $4.047\%$ & $98.662\%$
\\
    Bear-2 & $49$ & $4.572\%$ & $98.461\%$
 \\
	Bird & $83$ & $2.560\%$ & $95.914\%$
\\ 
    Bug & $137$ & $4.511\%$ & $98.274\%$
\\
    Bunny & $73$ & $5.771\%$ & $97.650\%$
\\
    Camel & $93$ & $5.556\%$ & $96.919\%$
\\
    Chair &	$115$ &	$2.947\%$ & $96.082\%$
\\
    Crab-1 & $93$ & $2.744\%$ & $97.874\%$
\\
    Crab-2 & $107$ & $2.825\%$ & $98.608\%$
\\
    Cup	& $243$ & $9.247\%$ & $95.411\%$
\\
    Dinosaur & $66$ & $2.672\%$ & $99.331\%$
\\ 
    Elephant & $94$ & $3.874\%$ & $98.079\%$
\\
    Fandisk	& $127$ & $4.067\%$ & $92.029\%$
\\
    Femur &	$26$ & $3.095\%$ & $97.536\%$
\\
    Fish & $43$ & $4.809\%$ & $88.587\%$
\\
    Giraffe	& $71$ & $4.208\%$ & $97.335\%$
\\ 
    Guitar & $71$ & $2.497\%$ & $96.804\%$
\\
    Hand & $47$ & $1.975\%$ & $95.143\%$
\\
    Human & $46$ & $2.105\%$ & $98.869\%$
\\
    Lifebuoy & $33$ & $3.324\%$ & $96.151\%$
\\
    Octopus & $74$ & $3.005\%$ & $97.989\%$
\\
    Plane & $44$ & $4.306\%$ & $83.389\%$
\\
    Pot	& $90$ & $5.278\%$ & $95.786\%$
\\
    Rocker & $112$ & $3.652\%$ & $96.944\%$
\\
    Seahorse & $60$ & $5.168\%$ & $96.144\%$
\\
    Spectacles & $97$ & $3.080\%$ & $96.075\%$
\\
    Spider & $54$ & $4.493\%$ & $95.987\%$
\\
    Vase & $118$ & $3.875\%$ & $95.710\%$
\\
\hline
{Average} & {-} & $3.938\%$ & $96.134\%$
 \\
\hline
\end{tabular}}
\end{center}
\vspace{-2mm}
\end{table}

\section{Complexity Analysis on Coverage Axis}
\label{supp:cat_slow}
\begin{figure}[t]
  \centering
  \begin{overpic}
[width=0.9\linewidth]{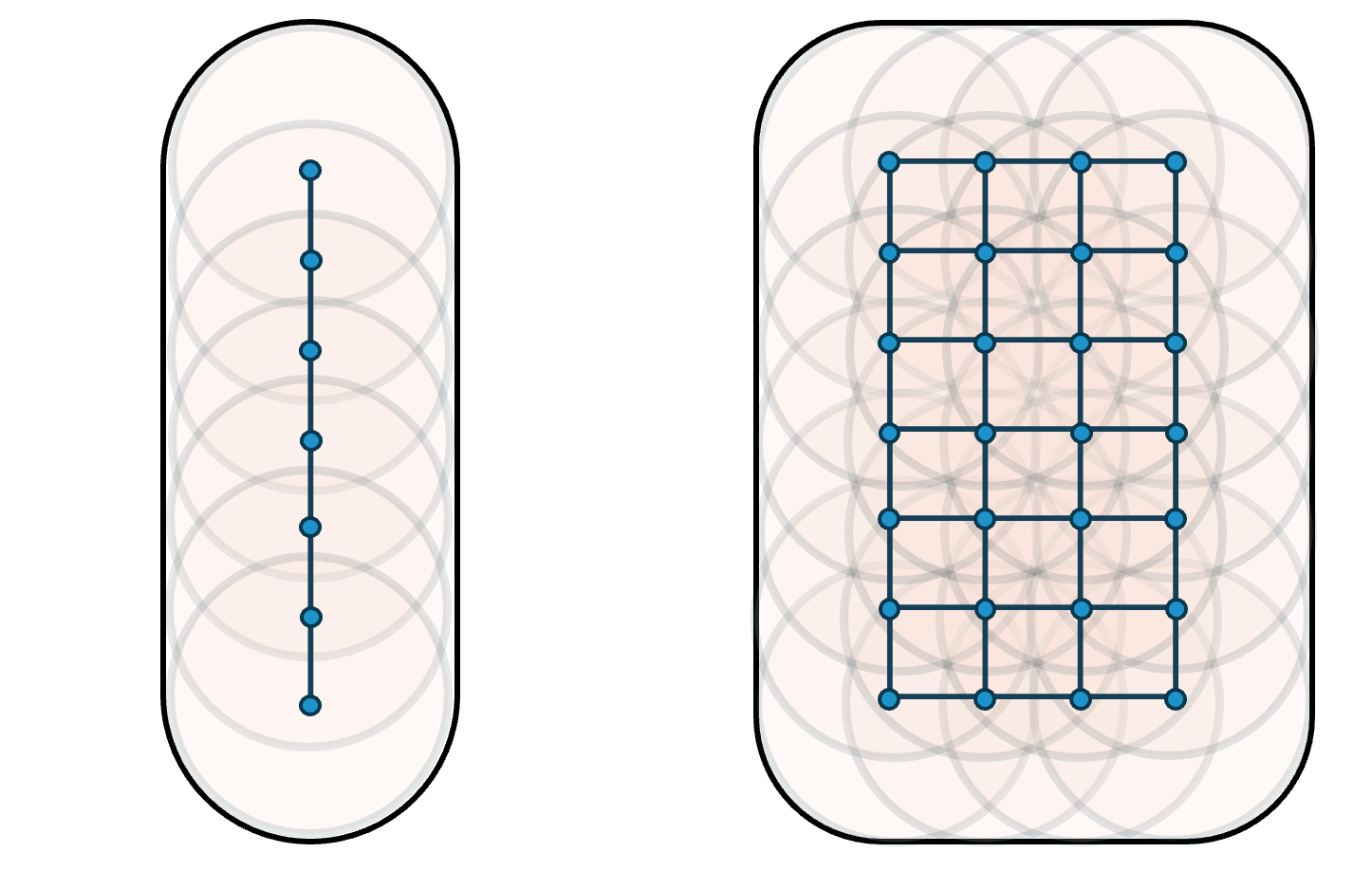}
\put(20.5,-3){(a)}
\put(73.5,-3){(e)}
\end{overpic}
    \caption{Distribution of randomly sampled candidate skeletal points in two different shapes. $(a)$ tubular structure. $(b)$ planar structure.}
    \label{fig:ca_flaw}
\end{figure}

\begin{table*}
\small
\begin{center}
\caption{Quantitative comparison on run time and shape approximation error between Coverage Axis and Coverage Axis++ when the number of surface samples is set as $5000$. The inputs are meshes. \textbf{Time}: \textit{Runtime measured in seconds.} $|V|$: \textit{The number of skeletal points.} \textit{Note we set the number of selected skeletal points of our method the same as the Coverage Axis~\cite{dou2022coverage} for a fair comparison.}} 
\label{table:more_surface}
\vspace{-2mm}
\setlength{\tabcolsep}{4mm}{
\begin{tabular}{p{1.3cm}|p{0.8cm}<{\centering}|cccc|cccc}
\hline
\multirow{3}{*}{Model} & \multirow{3}{*}{$|V|$ } & \multicolumn{4}{c|}{Coverage Axis}  & \multicolumn{4}{c}{Coverage Axis++}\\
  &  &  \textbf{Time} & $\overrightarrow{\epsilon}$ &  $\overleftarrow{\epsilon}$ &  $\overleftrightarrow{\epsilon}$ & \textbf{Time} &  $\overrightarrow{\epsilon}$ &  $\overleftarrow{\epsilon}$ &  $\overleftrightarrow{\epsilon}$\\
    \hline
Bear-1&	70&	115.2&	$3.651\%$ &	$4.958\%$ &	$4.958\%$ &	\toptime{$5.1$}&	$2.752\%$ &	$2.795\%$ &	\topac{$2.795\%$}\\
Bear-2&	38&	7.5&	$7.412\%$ &	$7.385\%$ &	$7.412\%$ &	\toptime{$2.5$}&	$4.552\%$ &	$5.997\%$ &	\topac{$5.997\%$}	\\
Bird&	101&	>1000&	$1.959\%$ &	$1.857\%$ &	\topac{$1.959\%$} &	\toptime{$9.8$} & $2.177\%$ &	$2.438\%$ &	$2.438\%$	\\
Bug&	145&	>1000&	$5.701\%$ &	$5.636\%$ &	$5.701\%$ &	\toptime{$12.2$}&	$3.074\%$ &	$3.005\%$ &	\topac{$3.074\%$} \\
Camel&	116&	25.6&	$2.488\%$ &	$2.238\%$ &	\topac{$2.488\%$} &	\toptime{$9.5$}&	$2.513\%$ &	$2.936\%$ &	$2.936\%$	\\
Chair&	137&	>1000&	$1.920\%$ &	$1.871\%$ &	\topac{$1.920\%$} &	\toptime{$13.6$} & $1.909\%$ &	$2.105\%$ &	$2.105\%$	\\
Crab-1&	92&	43.9&	$3.738\%$ &	$3.293\%$ &	$3.738\%$ &	\toptime{$9.5$}&	$2.577\%$ &	$2.738\%$ &	\topac{$2.738\%$}	\\
Crab-2&	83&	210.5&	$7.705\%$ &	$6.223\%$ &	$7.705\%$ &	\toptime{$8.9$}&	$3.168\%$ &	$3.212\%$ &	\topac{$3.212\%$}	\\
Cup & 286& >1000 & $9.176\%$ &	$5.026\%$ &	$9.176\%$ &	\toptime{$27.8$}&	$9.174\%$ &	$6.390\%$ &	\topac{$9.174\%$}	\\
Dog & 65 &	7.6 & $4.557\%$ &	$4.386\%$ &	$4.557\%$ &	\toptime{$5.0$} &	$2.642\%$ &	$2.876\%$ &	\topac{$2.876\%$} \\
Elephant&	89&	147.1&	$6.015\%$ &	$7.221\%$ &	$7.221\%$ &	\toptime{$8.5$}&	$3.168\%$ &	$3.925\%$ &	\topac{$3.925\%$}	\\
Fandisk&	210&	>1000&	$2.925\%$ &	$2.778\%$ &	\topac{$2.925\%$} &	\toptime{$23.7$}&	$2.613\%$ &	$3.057\%$ &	$3.057\%$	\\
Femur&	27&	13.6&	$2.842\%$ &	$2.782\%$ &	$2.842\%$ &	\toptime{$2.0$} & $2.035\%$ &	$2.035\%$ &	\topac{$2.035\%$}	\\
Fish&	72&	>1000&	$1.769\%$ &	$2.271\%$ &	\topac{$2.271\%$} &	\toptime{$5.7$} &	$1.994\%$ &	$2.722\%$ &	$2.722\%$	\\
Giraffe&	50&	25.9&	$5.841\%$ &	$5.853\%$ &	$5.853\%$ &	\toptime{$5.1$}&	$2.627\%$ &	$5.267\%$ &	\topac{$5.267\%$}\\
Guitar&	63&	>1000&	$2.040\%$ &	$4.312\%$ &	$4.312\%$ &	\toptime{$7.4$}&	$2.232\%$ &	$2.383\%$ &	\topac{$2.383\%$}	\\
Hand&	46&	131.4&	$2.874\%$ &	$2.607\%$ &	\topac{$2.874\%$} &	\toptime{$4.2$}&	$3.315\%$ &	$2.884\%$ &	$3.315\%$	\\
Horse&	56&	200.2&	$5.174\%$ &	$6.042\%$ &	$6.042\%$ &	\toptime{$4.6$}&	$3.608\%$ &	$5.246\%$ &	\topac{$5.246\%$}	\\
Lifebuoy&	38&	96.7&	$4.667\%$ &	$4.196\%$ &	$4.667\%$ &	\toptime{$2.8$} &	$2.52\%$ &	$3.307\%$ &	\topac{$3.307\%$}	\\
Neptune&	120&	206.8&	$2.115\%$ &	$5.751\%$ &	$5.751\%$ &	\toptime{$11.4$} &	$3.128\%$ &	$5.542\%$ &	\topac{$5.542\%$}	\\
Pig&	51&	16.6&	$8.535\%$ &	$8.575\%$ &	$8.575\%$ &	\toptime{$3.5$}&	$6.038\%$ &	$8.327\%$ &	\topac{$8.327\%$}	\\
Plane&	84&	>1000&	$1.708\%$ &	$2.275\%$ &	\topac{$2.275\%$} &	\toptime{$8.7$}&	$1.965\%$ &	$2.533\%$ &	$2.533\%$	\\
Pliers&	32&	13.5&	$2.072\%$ &	$1.925\%$ &	\topac{$2.072\%$} &	\toptime{$2.8$}&	$2.081\%$ &	$2.055\%$ &	$2.081\%$	\\
Pot&	152&	>1000&	$3.570\%$ &	$3.100\%$ &	\topac{$3.570\%$} &	\toptime{$14.4$}&	$3.504\%$ &	$3.690\%$ &	$3.690\%$	\\
Rapter&	74&	>1000&	$3.322\%$ &	$6.545\%$ &	\topac{$6.545\%$} &	\toptime{$6.2$}&	$3.322\%$ &	$6.634\%$ &	$6.634\%$	\\
Rocker&	130&	>1000&	$3.632\%$ &	$3.496\%$ &	\topac{$3.632\%$} &	\toptime{$14.4$}&	$3.777\%$ &	$2.790\%$ &	$3.777\%$	\\
Seahorse&	53&	>1000&	$2.763\%$ &	$11.156\%$ &	$11.156\%$ &	\toptime{$4.8$}&	$2.288\%$ &	$5.515\%$ &	\topac{$5.515\%$}	\\
Spectacles&	112&	>1000&	$1.982\%$ &	$1.952\%$ &	$1.982\%$ &	\toptime{$14.4$} &	$1.660\%$ &	$1.653\%$ &	\topac{$1.660\%$}	\\
Vase&	109&	226.1&	$5.260\%$ &	$6.587\%$ &	$6.587\%$ &	\toptime{$9.2$}&	$3.132\%$ &	$3.192\%$ &	\topac{$3.192\%$}	\\
Wine-glass&	335&	>1000&	$22.680\%$ &	$7.904\%$ &	$22.680\%$ &	\toptime{$34.6$}&	$21.310\%$ &	$8.549\%$ &	\topac{$21.310\%$}	\\
\hline
{Average} & {-} & 516.3 & 4.670\% & 4.673\% & 5.448\% & \toptime{$9.7$} & 3.695\% & 3.860\% & \topac{$4.429\%$}
 \\
\hline
\end{tabular}}
\end{center}
\vspace{-2mm}
\end{table*}

Next, we provide more analysis of the complexity of the Coverage Axis. 
From Tab.~1 of the main paper, we observe that the running times of the Coverage Axis for certain models (such as Bird, Chair, Guitar, and Spectacles) are extremely long, indicating that solving SCP for these models is particularly challenging. Note that a common characteristic of these shapes is that they all have planar shapes. We illustrate why solving SCP for models with planar shapes is difficult.

Given an arbitrary shape, under the meaning of medial axis transform, it can be divided into two types of shapes: \textit{tubular} and \textit{planar} structures, as depicted in Fig.~\ref{fig:ca_flaw}. Without loss of generality, we assume that the size and sample density of the two shapes are the same. In such cases, if the tubular shape contains $q$ points, the planar shape would have $q^2$ points. Recall that the SCP amounts to the following $0-1$ integer programming:
\begin{equation}
\begin{split}  
&\min \left\|\mathbf{v}\right\|_2, \\  
&s.t.\  \begin{array}{c}  
\mathbf{Dv} \ge \mathds{1},\\
\end{array}  
\end{split}  
\label{eq:core_op}
\end{equation} 
where $\mathds{1}$ is the vector of all ones, $\ge$ is applied element-wise, and $\mathbf{v} \in \{0,1\}^{n\times1}$ is a decision vector, with its $i$-th element $v_i\in\{0,1\}$ indicating if the candidate skeletal point $p_i$ is selected.
Consequently, for the tubular shape, there are $2^q$ possible combinations of $v$, while for the planar shape, there are $2^{q^2}$ possible combinations. As a result, solving the SCP becomes significantly more challenging when the model involves planar shapes due to the exponential increase in the number of possible combinations. In comparison, Coverage Axis++ sidesteps solving an optimization problem with high computation complexity by designing a simple yet effective heuristic algorithm, taking coverage and uniformity into consideration, which thus achieves significant acceleration.

\section{Skeletal Point Connectivity}
\label{supp_sec:why_mesh}
\begin{figure}[t]
\vspace{5mm}
  \centering
  \begin{overpic}
[width=\linewidth]{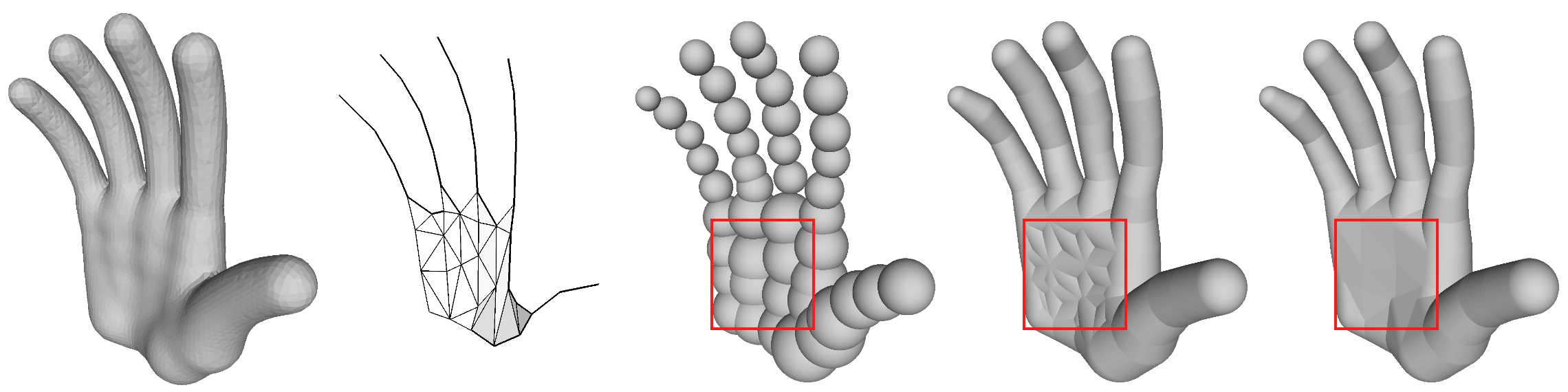}
\put(9,-4.5){(a)}
\put(29,-4.5){(b)}
\put(49,-4.5){(c)}
\put(69,-4.5){(d)}
\put(89,-4.5){(e)}
\end{overpic}
\vspace{-0.5mm}
\caption{The medial mesh part among the skeletal points in the MAT enables interpolation between medial axis balls, resulting in higher accuracy shape reconstruction (e). In contrast, reconstruction using only skeletal points (c) or without a medial mesh connection (d) typically leads to larger approximation errors.}
    \label{fig:discussion_connection_recon}
\end{figure}

\begin{figure}[t]
\vspace{5mm}

  \centering
  \begin{overpic}
[width=\linewidth]{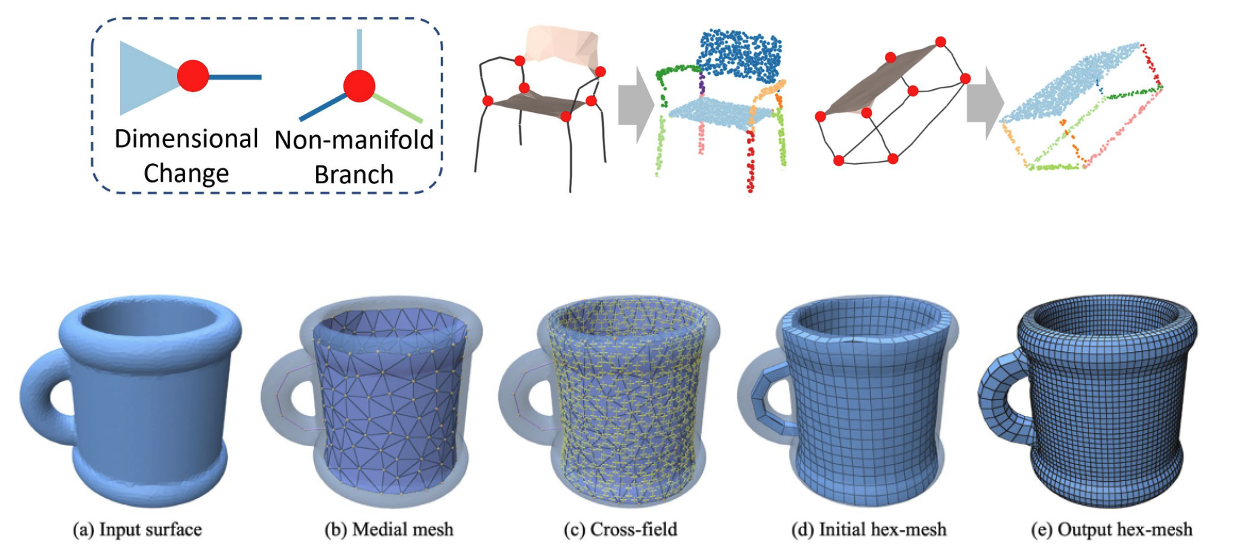}
\put(40,24){(a)~\cite{lin2021point2skeleton}}
\put(39,-4){(b)~\cite{zhang2024medial}}
\end{overpic}
\vspace{-1mm}
\caption{(a) The medial axis transform (MAT) includes a medial surface. Unsupervised shape segmentation methods show that detecting dimensional changes and non-manifold branches on the medial mesh effectively aids segmentation ~\cite{lin2020seg}. (b) In the application of all-hex meshing~\cite{zhang2024medial}, the medial axis helps with computing the cross-field to guide the generation of the hexahedral mesh during the optimization.}
    \label{fig:discussion_connection_apps}
\end{figure}

\ZY{In this Section, we discuss the necessity of mesh connectivity in the medial axis for various downstream applications. 1) Reconstruction: The connectivity of meshes and edges in the Medial Axis Transform (MAT) is important for shape reconstruction. This is because the connectivity allows for interpolation between medial axis balls, a.k.a., slab meshes, thereby achieving relatively higher accuracy in shape reconstruction. In this process, the connectivity between medial axis points is key to establishing this interpolation. If we disregard this connectivity, the results will exhibit obvious errors; See Fig.~\ref{fig:discussion_connection_recon}. 2) Shape Segmentation: As MAT contains a medial surface, previous unsupervised structural decomposition/shape segmentation~\cite{lin2020seg,lin2021point2skeleton} has shown that detecting dimensional changes and non-manifold branches on the medial mesh can be effective for the shape segmentation, e.g., use the information from manifold branch and dimensional change for segmentation; See Fig.~\ref{fig:discussion_connection_apps} (a). 3) Hexmeshing: Based on the shape description ability of medial mesh, one could produce all-hex meshing using topological and geometrical information of the medial mesh, which is more general than previous methods based on curve-skeleton as pointed out by~\cite{zhang2024medial}; See Fig.~\ref{fig:discussion_connection_apps} (b).}

